\documentclass[sigconf]{acmart}

\AtBeginDocument{%
	}

\copyrightyear{2023} 
\acmYear{2023} 
\setcopyright{rightsretained} 
\acmConference[CIKM '23]{Proceedings of the 32nd ACM International Conference on Information and Knowledge Management}{October 21--25, 2023}{Birmingham, United Kingdom}
\acmBooktitle{Proceedings of the 32nd ACM International Conference on Information and Knowledge Management (CIKM '23), October 21--25, 2023, Birmingham, United Kingdom}\acmDOI{10.1145/3583780.3615036}
\acmISBN{979-8-4007-0124-5/23/10}

\begin{document}
	\title{Reveal the Unknown: Out-of-Knowledge-Base Mention Discovery with Entity Linking} 
	
	\author{Hang Dong}
	\affiliation{%
		\institution{University of Oxford}
		\country{}}
	\email{hang.dong@cs.ox.ac.uk}
	
	\author{Jiaoyan Chen}
	\affiliation{%
		\institution{The University of Manchester \& University of Oxford}
		\country{}}
	\email{jiaoyan.chen@manchester.ac.uk}
	
	\author{Yuan He}
	\affiliation{%
		\institution{University of Oxford}
		\country{}}
	\email{yuan.he@cs.ox.ac.uk}

	\author{Yinan Liu}
	\affiliation{%
		\institution{Nankai University}
		\country{}}
	\email{liuyn@mail.nankai.edu.cn}
	
	\author{Ian Horrocks}
	\affiliation{%
		\institution{University of Oxford}
		\country{}}
	\email{ian.horrocks@cs.ox.ac.uk}
	
	\renewcommand{\shortauthors}{Hang Dong, Jiaoyan Chen, Yuan He, Yinan Liu, \& Ian Horrocks}
	
	\begin{abstract}
		Discovering entity mentions that are out of a Knowledge Base (KB) from texts plays a critical role in KB maintenance, but has not yet been fully explored. The current methods are mostly limited to the simple threshold-based approach and feature-based classification, and the datasets for evaluation are relatively rare. We propose BLINKout, a new BERT-based Entity Linking (EL) method which can identify mentions that do not have corresponding KB entities by matching them to a special NIL entity. To better utilize BERT, we propose new techniques including NIL entity representation and classification, with synonym enhancement. We also apply KB Pruning and Versioning strategies to automatically construct out-of-KB datasets from common in-KB EL datasets. Results on five datasets of clinical notes, biomedical publications, and Wikipedia articles in various domains show the advantages of BLINKout over existing methods to identify out-of-KB mentions for the medical ontologies, UMLS, SNOMED CT, and the general KB, WikiData.\footnote{The implementation, dataset construction scripts, and datasets are publicly available at \url{https://github.com/KRR-Oxford/BLINKout} and \url{https://zenodo.org/record/8228371}.}
	\end{abstract}

	
	
	\maketitle
	
	\section{Introduction}
	Knowledge Bases (KBs) are widely used for representing entities and facts about the world with reasoning supported. 
	KBs are inherently incomplete. New entities are constantly emerging, for example, from late 2020 to early 2022, a new variant of SARS-CoV-2 emerged every few months \cite{who2022variants}. Existing KBs may thus also inevitably miss entities, for example, ``Curry-Jones syndrome'' \cite{twigg2016recurrent} was not added to SNOMED CT \cite{donnelly2006snomed} until 2017. Delays in incorporating these entities into the KB may result in failure to cataloguing, searching, and reasoning with them. Automated discovery of mentions of new entities from common resources such as texts can support the maintenance of KBs.
	
	Updating KBs with entities from texts is highly relevant to \textit{Entity Linking} (EL), which is to match mentions in texts to entities in a KB \cite{poibeau_entity_2013,shen2014entity}. Current works on EL, however, often assume all the target mentions have corresponding entities in the KB and ignore mentions that have no corresponding entities \cite{wu2020blink,basaldella2020cometa}. The latter are sometimes called \textit{out-of-KB mentions} and are matched with a \textit{NIL entity}. To maintain a KB, it is required to discover out-of-KB mentions which can be further processed as new KB entities.\footnote{In this work, we focus on identifying out-of-KB \textit{mentions}, and leave the canonicalisation and placement of the new entities in the KB for future work.}
	
	As NIL is not described in the KB, it is hard to obtain their lexical or embedding representations, and consequently, it is hard to predict them directly. One idea is setting a threshold towards the mention-to-entity matching score: a mention is regarded as NIL if its scores to all the KB entities are below the threshold \cite{bunescu-pasca-2006}. Another idea is to \textit{classify} a mention into NIL or in-KB based on a set of features regarding the mention and its top-\textit{k} entity candidates \cite{wu2016features}. There are also some other studies attempting to represent the NIL entity with key phrases or features based on external corpora \cite{Hoffart2014} or manual effort \cite{Hoffart2016}. 
	
	Recently, neural EL has shown good performance by applying pre-trained language models (LMs), e.g., BERT \cite{devlin-etal-2019-bert}, to represent texts and entities \cite{wu2020blink}. In contrast, such LM-based methods have been rarely applied for out-of-KB mentions, as reviewed in \cite{shen2021eldeep,sevgili_neural_2022}. Traditional, threshold-based approaches in combination with deep learning are still the state-of-the-art \cite{ji2020bert,chen2021lightweight}. There is a lack of studies to seamlessly integrate out-of-KB mention discovery with pre-trained LM-based approaches like BLINK \cite{wu2020blink}. Also, entities are likely to have various surface forms or synonyms \cite{dsouza-ng-2015-sieve-based,chen2021lightweight}. This entity variety suggests enhancing synonyms for EL and may help differentiate between in-KB and out-of-KB entities.
	
	Besides the shortage of methodology research, benchmarking datasets considering out-of-KB mention discovery are also relatively rare compared to in-KB EL. Most EL datasets assume that the KB is complete and do not include out-of-KB entities, e.g., the MedMentions dataset \cite{mohan2019medmentions} that links mentions to concepts in UMLS \cite{Bodenreider2004umls}. The most recent large-scale dataset is NILK \cite{Iurshina2022nilk}, which synthesizes NIL entities from the entity gap between the newer version of Wikidata (2021) and the older version (2017) to enrich the older KB. Also, in the biomedical domain, the main out-of-KB EL dataset is Share/CLEF 2013 \cite{suominen2013shareclef}, with NIL mentions manually identified for training and evaluation. The only two other datasets from the other domains also rely on manual annotations, e.g., historical newspapers \cite{ehrmann2020extended} and news in microposts \cite{rizzo2017NEEL}, which require domain experts' substantial effort. Also, all the previous studies only focus on a single strategy (either manual effort or KB versioning) to construct EL datasets with NIL labels.
	
	In this work, (i) we define the out-of-KB mention discovery problem and propose a method named BLINKout, based on the BERT-based EL method, BLINK \cite{wu2020blink}, where new techniques including NIL entity representation \& classification and synonym enhancement are developed and applied; (ii) we summarize and apply strategies to automatically construct out-of-KB mention discovery benchmarks from an in-KB EL dataset by pruning or using older versions of the linked ontology, besides manual labelling. Five out-of-KB mention discovery datasets are selected and constructed using three data strategies (i.e., manual labelling, KB pruning, KB versioning), with texts of clinical notes, biomedical publications and Wikipedia articles, two medical ontologies (UMLS and SNOMED CT), and one general purpose KB (WikiData) covering various domains.
	
	Experimental results on the datasets show the advantage of the BLINKout approach for out-of-KB mention discovery, with comparison to rule-based, threshold-based, and feature-based baselines and ablation studies, up to nearly 40\% improvement of out-of-KB $F_1$ score compared to the second best baseline.

	\section{Preliminaries}
	\subsection{Problem Definition}
	Given a KB containing a set of entities $E$ and a text corpus in which a set of mentions $M$ are identified in advance, \textit{Entity Linking} (EL, a.k.a. Entity Disambiguation or Entity Normalisation) is to map each mention $m$ in $M$ to its corresponding entity $e$ in $E$ \cite{shen2014entity}. Due to the limited knowledge coverage of the KB, it is possible that there is no entity in $E$ that can be matched with a given mention. Thus we have the following more general problem of \textbf{out-of-KB mention discovery}, formulated as an extended EL task that focuses on identifying out-of-KB mentions along with linking in-KB mentions. 
	
	Given a text corpus with a set of identified mentions (each in a context window in a document) $D_M$ and a KB with entities $E$, \textit{out-of-KB mention discovery} is to develop a function $f$ such that each mention $d_m$ in $D_M$ is mapped to an item $f(d_m) \in E \cup \{\text{NIL}\}$, where \text{NIL} is a special entity denoting that there are no entities in $E$ that can be matched with $d_m$. In this study, we focus on both ontologies (e.g., SNOMED-CT) and general KBs (e.g., WikiData). Ontologies are a common type of KB that is often defined as a shared, explicit specification of a conceptualisation of a domain \cite{GRUBER1995907}, and we consider an ontology's classes (or concepts) as its entities for the linking. 
	Note that each entity may have definitions and synonyms that can be utilized for out-of-KB mention discovery. 
	
	\subsection{BERT-based Entity Linking}\label{bert-based-el}
	We summarize the BERT-based Entity Linking methods (e.g., BLINK \cite{wu2020blink}) below. They usually have two stages: \textit{candidate creation} and \textit{candidate ranking} \cite{poibeau_entity_2013,ji2020bert,
		wu2020blink}. In candidate creation, the approaches aim to narrow down the vast number of entities into a manageable subset (e.g., tens or hundreds), and in candidate ranking, the approaches aim to rank the candidate entities of each mention according to the probability that they match the given mention.
	
	\textbf{Candidate Creation with Bi-encoder}. The bi-encoder fine-tunes two BERT models $BERT_m$ and $BERT_e$ to embed mentions and entities, resp., into a dense embedding space, so that given a new mention, the nearest candidates can be easily retrieved. For a mention $d_{m}$ and an entity $d_e$, their embeddings can be accessed as
	\begin{equation}
		v_m = \text{red}(BERT_m(d_{m})); \hspace{0.3cm} v_e = \text{red}(BERT_e(d_{e}))
	\end{equation}
	where $\text{red}(\cdot)$ denotes a function for extracting the vector representation of a textual sequence from a BERT model using its last layer representation of the \texttt{[CLS]} token. 
	
	Mentions and entities can be fed into the BERT models in different ways. In the classic work \cite{wu2020blink}, a mention is fed into $BERT_m$ as \texttt{[CLS] ctxt$_l$ [M$_s$] mention [M$_e$] ctxt$_r$ [SEP]}, where ctxt$_l$ and ctxt$_r$ are the left and right contexts of the mention in the document, resp.; and an entity (with name and definition) is fed into $BERT_e$ as \texttt{[CLS] name [ENT] definition [SEP]}. Note that \texttt{[M$_s$]}, \texttt{[M$_e$]}, and \texttt{[ENT]} are special tokens for separation. 
	
	With these embeddings, the \textbf{score} which measures the similarity between a mention and an entity can be calculated, e.g., by their dot-product defined as $s(m,e) = v_m \cdot v_e$. The scores are further used to generate top-$k$ candidates for each mention.
	
	\begin{figure*}[t!]
		\includegraphics[width=0.95\textwidth]{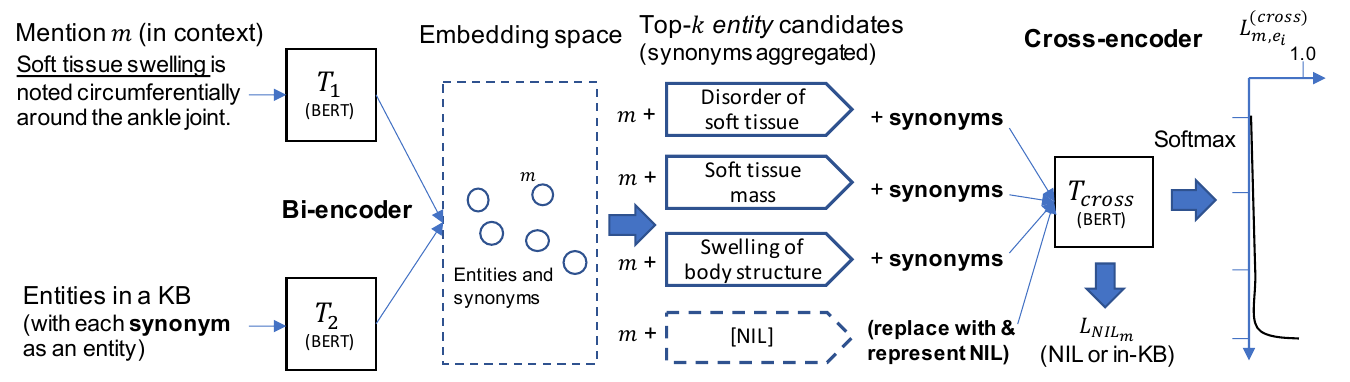}
		\caption{BLINKout architecture for out-of-KB mention discovery, adapting BERT-based Entity Linking \cite{wu2020blink}: \textbf{bi-encoder} encodes separately the mention $m$ in a context and the entities (each synonym as an entity) into a dense embedding space; \textbf{cross-encoder} classifies the most relevant entity candidate (with synonym concatenated), with \textbf{NIL Entity Representation \& Classification} that appends a \texttt{[NIL]} special token to replace the last candidate (if NIL was not predicted by the bi-encoder), jointly learned with $L_{NIL}$ to classify whether the mention is out-of-KB.}\label{BLINKout-architecture}
		\vspace{-3mm}
	\end{figure*} 
	
	Bi-encoder can be trained with a loss function to make each mention close to its matched entity in the embedding space, but far away from the other entities within the same batch. This can be realised with the max-margin triplet loss \cite{reimers-gurevych-2019-sentence,basaldella2020cometa} described below\footnote{The bi-encoder is originally trained using a cross-entropy loss \cite{wu2020blink} (or a non-temperatured contrastive loss \textit{cf.} \citet{gao-etal-2021-simcse}) for a similar intuition. We found that the triplet loss led to a more stable convergence than the cross-entropy loss in \cite{wu2020blink}, with a similar performance to the latter, in the experiments.}, where $\alpha$ is a margin of small value (e.g., 0.2) and $[x]_+$ denotes $\text{max}(x,0)$, for each mention to its gold entity (the $i$-th) in a batch.
	\begin{equation}\label{loss-bi}
		L_{m_i,e_i} = \sum_{j \neq i}{[\alpha-s(m_i,e_{i})+s(m_i,e_{j})]_{+}}
	\end{equation}
	
	\textbf{Candidate Ranking with Cross-encoder}. Given that the bi-encoder can only coarsely identify top-k candidates but may not correctly rank them among a large number of KB entities \cite{wu2020blink}, we use the cross-encoder to make the candidate ranking. The cross-encoder is fed with information of both a mention and its top-$k$ entities from the bi-encoder, and performs a multi-class classification, i.e., the entity that is more likely to be matched when the mention is predicted with a higher score. 
	
	Each mention $m$ is concatenated with each of its top-$k$ entities following the same input format as in the bi-encoder. The concatenation of the input for the mention and the entity $e$ (without the \texttt{[SEP]} and \texttt{[CLS]} tokens in between) is denoted as $d_{m,e}$, i.e., \texttt{[CLS] ctxt$_l$ [M$_s$] mention [M$_e$] ctxt$_r$ [SEP] name [ENT] definition [SEP]}. The input $d_{m,e}$ is fed into the cross-encoder, which is composed of a BERT model and a linear layer, for a score output $s^{(cross)}_{m,e}$. The vector $w$ constitutes the parameters to be learned in the model.
	\begin{equation}\label{cross-enc}
		s^{(cross)}_{m,e} = v^{(cross)}_{m,e}w; \hspace{0.3cm} v^{(cross)}_{m,e} = \text{red}(BERT_{cross}(d_{m,e}))
	\end{equation}
	
	The cross-encoder is learned with the following cross-entropy loss with softmax activation, where $E_{cand}$ denotes the indices of the top-$k$ candidate entities of $m$, and $i$ is the index of the gold entity.
	\begin{equation}\label{loss-cross-enc}
		L^{(cross)}_{m,e_i} = -\text{log}(\frac{\text{exp}(s^{(cross)}_{m,e_i})}{\sum_{j \in E_{cand}} \text{exp}(s^{(cross)}_{m,e_j})})
	\end{equation}
	
	\section{BLINKout}
	\label{NIL-ent-methods}
	The BERT-based EL \cite{wu2020blink} described previously in Section \ref{bert-based-el} does not consider the NIL entity. We present NIL entity representation \& classification, an approach for out-of-KB mention discovery that adapts the cross-encoder, applicable either with or without training, and discuss the techniques to represent the NIL entity with LMs. Then, we propose methods to enhance both bi-encoder and cross-encoder with synonyms or variants which are prevalent for entities.
	
	\subsection{NIL Entity Classification}
	For the classification approach, we ensure that the NIL entity is within the top-$k$ candidates by replacing the last candidate with NIL. This gives a chance for the cross-encoder with $L^{(cross)}$ to potentially classify NIL as the top entity for the mention, as inspired by \cite{poibeau_entity_2013}. The textual representation of NIL, encoded with BERT-like LMs, will affect the performance of classification; we discuss different NIL entity representations in \ref{NIL-rep}.
	
	\textbf{Joint Training for NIL Classification}. We add a joint loss to learn to verify whether a mention is NIL or in-KB, conditioned on the mention representation, similar to \cite{gu2021read}. The sigmoid function $\sigma$ with a linear layer is used to form a probability score, $s_{isNIL_m}$, given the mention representation. The vector $w_{NIL}$ constitutes the model parameters, learned with the binary cross-entropy loss below.
	\begin{equation}\label{cross-enc-NIL}
		s_{isNIL_m} = \sigma (v^{(cross)}_{m}w_{NIL}); \hspace{0.1cm} v^{(cross)}_{m} = \text{red}(BERT_{cross}(d_{m}))\\
	\end{equation}
	\begin{equation}\label{loss-nil}
		L_{NIL_m} = -y_i\text{log}(s_{isNIL_m}) + (1-y_i)\text{log}(1-s_{isNIL_m})
	\end{equation}
	
	The overall loss with joint training for the cross-encoder is described below. The best value of $\lambda_{NIL}$ varied from 0.01 to 0.25 across the datasets based on parameter tuning with the validation set.
	\begin{equation}\label{loss-overall}
		L_m = L^{(cross)}_{m,e_i} + \lambda_{NIL} * L_{NIL_m}
	\end{equation}
	
	\subsection{NIL Entity Representation}
	\label{NIL-rep}
	
	The NIL entity does not have a native textual representation in the KB. A better representation of NIL will help the LM to discriminate NIL from in-KB entities. This representation can be either static, fixed (or unsupervised) or dynamic, fine-tuned (or supervised with NIL mentions) in the LM.
	
	We represent NIL as a special token \texttt{[NIL]} by taking advantage of the tokenizer in a BERT-like LM. This assigns special semantics to the NIL entity so it is not confused with the names of other entities in the KB. Also, the continuous representation of \texttt{[NIL]} can be further fine-tuned with the LM. A more naive representation is ``NIL'', with the definition of ``It is a NIL option.'', used in a previous study \cite{gu2021read}; we also replace NIL with \texttt{[NIL]} in the definition. Similar to in-KB entities, we add the \texttt{[ENT]} special token between the name and the definition. A list of NIL entity representations and their results are presented in Table \ref{NIL-rep-results}. Our final approach uses the dynamic, fine-tuned \texttt{[NIL]} representation that leverages the labelled out-of-KB mentions in the training data.
	
	\subsection{Synonym Enhancement}
	The original BLINK \citep{wu2020blink} focuses on Wikipedia texts and does not consider synonyms. Synonyms are prevalent for real-world entities (see data statistics in Table \ref{data-stats}), e.g., entity C0428977 in the UMLS has the name of ``Bradycardia'' and some synonyms such as ``Slow heart beat'' and ``Heart rare slow''. When in-KB entities are better represented with synonyms, the out-of-KB mentions can be more precisely identified by the LM. We thus enhance the bi-encoder and the cross-encoder with synonyms, with two different approaches, resp.: (i) augmentation of each synonym as an entity in the bi-encoder, for its thorough training, and (ii) concatenation with \texttt{[SYN]} special token in the cross-encoder, for its efficient training.
	
	\textbf{Synonym Augmentation in Bi-encoder.} We use synonyms for data augmentation to enhance the training. Each synonym is treated as a separate entity to be matched to the mention. This can significantly augment the size of the training data. After the scoring, we aggregate the entities and synonyms into top-$k$ unique entity candidates by setting the score of an entity as its highest score among all its variations.
	
	\textbf{Synonym Concatenation in Cross-encoder.} It is inefficient and infeasible to use synonym augmentation in the cross-encoder, as the number of classes can be significantly increased (e.g., by around 3-4 times for the UMLS and SNOMED CT subset) and unstable or non-fixed. Thus, 
	we model each entity candidate $e$ with the concatenation of its synonyms, separated by the \texttt{[SYN]} special tokens, i.e., \texttt{[CLS] name [ENT] synonym\_1 [SYN] synonym\_2 ... synonym\_n [SYN] definition}.\footnote{For bi-encoder, it is also possible to use synonym concatenation, while the proposed synonym augmentation performed better on Share/CLEF 2013 in the experiments.} 
	This keeps the number of entities to classify to $k$ instead of treating each synonym as an entity, and thus significantly reduces the computation while still making full use of the synonyms.
	
	Finally, we use \textbf{BLINKout} to refer to the approach that integrates BERT-based EL with synonym enhancement and fine-tuned NIL entity representation for classification. 
	
	\section{Dataset Construction}
	
	\begin{table*}[t!]
		\small
		\center
		\renewcommand{\arraystretch}{0.79} 
		\begin{tabular}{l|l|ll|ll}
			\toprule
			& ShARe/CLEF 2013 & MM-pruned-0.1  & MM-pruned-0.2  & MM-2014AB & NILK-sample \\
			& train / test & train / valid / test & train / valid / test & train / valid / test & train / valid / test  \\
			\midrule
			Out-of-KB mention creation & Manual Labelling & KB Pruning & KB Pruning & KB Versioning & KB Versioning \\
			Domain or document type & Clinical notes & PubMed abstracts & PubMed abstracts & PubMed abstracts & Wikipedia articles \\
			\# of docs        & 199 / 99     & 2,635 / 878 / 879     & 2,635 / 878 / 879     & 2,635 / 878 / 879 & 81,334 / 10,587 / 10,528 \\ 
			\# of mentions        & 5,816 / 5,351 & 6,201 / 2,121 / 2,000   & 6,201 / 2,121 / 2,000   & 6,181 / 2,112 / 1,988 & 86,379 / 10,688 / 10,613 \\ 
			\# of out-of-KB mentions   & 1,639 / 1,723  & 456 / 201 / 155      & 1,031 / 405 / 415     & 307 / 82 / 103 & 1,312 / 154 / 158 \\ 
			\% of out-of-KB mentions  & 28.2\% / 32.3\%  & 7.3\% / 9.5\% / 7.8\% & 16.6\% / 19.1\% / 20.8\% & 5.0\% / 3.9\% / 5.2\% & 1.5\% / 1.4\% / 1.5\% \\ 
			\# of in-KB mentions & 4,109 / 3,604  & 5,745 / 1,920 / 1,845   & 5,170 / 1,716 / 1,585   & 5,874 / 2,030 / 1,885 & 85,067 / 10,534 / 10,455 \\
			\% mentions w/ zero-shot entities & - / 11.6\% & - / 24.8\% / 25.3\% & - / 23.2\% / 22.0\% & - / 27.8\% / 28.0\% & - / 98.6\% / 98.5\%\\ 
			\midrule
			\# of entities in KB      & 88,150     & 35,392           & 31,460           & 35,398      & 79,412 \\ 
			\# of entities \& synonyms in KB       & 288,490    & 126,188          & 112,097          & 124,132    & 121,191 \\
			\% of entities having synonyms        & 99.0\% & 99.4\% & 99.4\% & 99.3\% & 25.5\% \\
			Ave \# syns per entity (having syns)          & 2.3 (2.3)  & 2.6 (2.6) & 2.6 (2.6) & 2.5 (2.5) & 0.5 (2.1) \\
			\bottomrule
		\end{tabular}
		\caption{Statistics of out-of-KB linking datasets. Slashes separate the statistics of training, validation, and testing sets. ``MM'' denotes MedMentions; pruned-0.1 and pruned-0.2 denote the percentage of pruned concepts in the ontology; ``2014AB'' refers to the older version of the UMLS applied; ``NILK-sample'' denotes the random sample of the NILK dataset that enriches WikiData.}\label{data-stats}
		\vspace{-3mm}
	\end{table*}
	
	There are different strategies to construct NIL-enhanced (or NIL-labelled) EL datasets, which contain out-of-KB mentions labelled with NIL (i.e., each mention is linked to either an entity in $E$ or NIL). One straightforward way is \textbf{Manual Labelling}. We adopt one main manually NIL-labelled dataset in the medical domain, ShARe/CLEF 2013, which consists of clinical notes (discharge summaries and electrocardiogram, echocardiogram, and radiology reports) in the version 2.5 of the MIMIC-II dataset \cite{suominen2013shareclef}. The target ontology is SNOMED CT (represented with their mapped UMLS CUI) refined with the Disorder semantic group of 10 semantic types in the UMLS, defined in the annotation guideline \cite{shareclef2013guideline}. We use UMLS 2012AB following \citet{ji2020bert}. Around 30\% of the mentions are out-of-KB.\footnote{The out-of-KB mentions can be out of the Disorder semantic group but belong to another semantic type (e.g., ``[Thick] bronchial secretions'' only as body substance; ``soft tissue swelling'' and ``vomiting coffee grounds'' only as findings) \cite{shareclef2013guideline}.}
	
	Manual labelling costs much labour. We thus apply two automatic strategies, KB Pruning and KB Versioning, to synthesize out-of-KB mentions within normal EL datasets. 
	
	\textbf{KB Pruning}. We randomly sample a portion (e.g., 10\% and 20\%) of the entities in the target KB and remove them from the KB. To preserve the hierarchies (formed by subsumption relations, a.k.a. ``subclass of'' or ''isA'' relations) in an ontology, we link the parent and child of each removed entity as in \citet{yuan2022resource,he2023deeponto}\footnote{We used the pruning function implemented in the Python library, DeepOnto \cite{he2023deeponto}, see \url{https://krr-oxford.github.io/DeepOnto/deeponto/onto/pruning/}.} and this forms a new ontology. Mentions that are originally linked to the removed entities are labelled by NIL.
	
\begin{figure}[t]
		\includegraphics[width=0.45\textwidth]{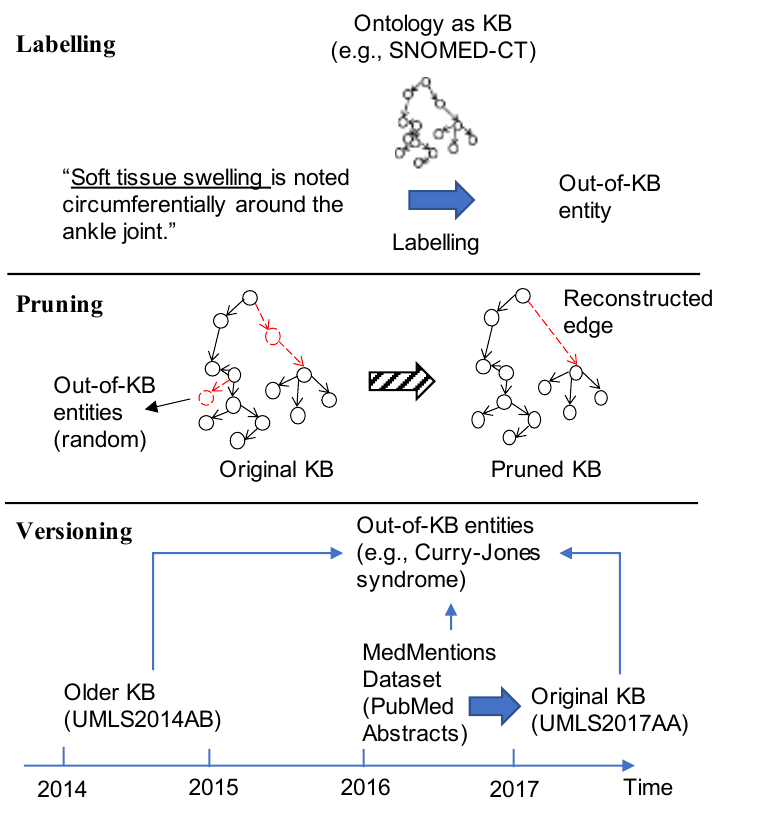}
		\caption{Strategies for NIL-enhanced EL dataset construction: Manual Labelling, KB Pruning, and KB Versioning. The specified UMLS ontologies are the subset linked to SNOMED CT under the semantic type of T047, ``Disease or Syndrome''.
		}\label{out-of-KB-ent-illustration} 
		\vspace{-3mm}
	\end{figure}

	\textbf{KB Versioning}. We also consider replacing the current version of the KB with an older version, so that mentions linked to entities which are in the newer version but not in the older version become out-of-KB. Using the MedMentions dataset \cite{mohan2019medmentions} as an example (see Figure \ref{out-of-KB-ent-illustration}), UMLS2017AA was used to exhaustively annotate entities in publication titles and abstracts in 2016; we chose its older version of UMLS2014AB (released in Nov 2014). We narrowed the UMLS semantic type to T047, \textit{Disease or Syndrome}, and selected the source as ``SNOMEDCT\_US'' (similar to ShARe/CLEF 2013). We gathered the entities that are in UMLS2017AA but not in the earlier version\footnote{We further filtered out those that were merged to a concept in the previous version of the ontology (as defined in Retired CUI Mapping, MRCUI.RRF in the UMLS), representing around 6\% of all newly added concepts.}. Mentions labelled with these new entities become out-of-KB\footnote{Examples include ``CJS'' or ``Curry-Jones syndrome'' (C0795915), ``CKD'' or ``Chronic Kidney Diseases'' (C1561643), and ``Pandemic influenza'' (C4304383), which were added to the UMLS between Nov 2014 and May 2017 under the semantic type T047, Disease or Syndrome.}. 
	
	We also adapted a recent general domain dataset constructed using the KB versioning approach. The recent NILK dataset \cite{Iurshina2022nilk} is created on the general, multi-domain KB, WikiData \cite{vrandecic2014wikidata}. The WikiData\_2021 version is used to supplement entities in WikiData\_2017. In-text hyperlinks (each link to another Wikipedia article) in the Wikipedia 2017 dump\footnote{\url{https://archive.org/download/enwiki-20170220/enwiki-20170220-pages-articles.xml.bz2}.} are used as the source of mentions; the hyperlinked Wikipedia articles are mapped to WikiData ID to generate in-KB (in WikiData\_2017) and out-of-KB entities (not in WikiData\_2017 but in WikiData\_2021) for the mentions \cite{Iurshina2022nilk}. We benchmark with a proportional random sample of 0.1\% of the mentions for each of the original data splits, given that the whole NILK dataset is huge (over 107M mentions each to be linked to one of 14.6M English-described entities in the full WikiData dump). We also select entities in the WikiData KB which have at least one mention linked after the random sampling of the mentions. This results in data sample and entity sizes comparable to the other four datasets and keeps the same proportion of NIL mentions as the original data: around 1.5\%, much lower than the other datasets.
	
	The advantage of KB Pruning is that it allows us to simulate the creation of out-of-KB entities and control their percentage, while KB Versioning creates datasets that are closer to real-life situations where new concepts emerge over time. 
	
	Statistics of the datasets are shown in Table \ref{data-stats}, including the counts of mentions and documents, and the percentages of out-of-KB mentions, zero-shot mentions, and entities with synonyms. The KB Versioning approach resulted in a much lower percentage of out-of-KB mentions. Also, NILK-sample is a zero-shot EL dataset (from its original construction in \cite{Iurshina2022nilk}), i.e., no overlapping entities (except NIL) among the training, validation, and testing sets, while in the other datasets most of the entities are seen in training. We also note that only about a quarter (25.5\%) of the sampled WikiData entities are associated with at least one synonym, compared to nearly all ($>$99.0\%) entities in the subset of UMLS and SNOMED-CT having synonym(s). The average number of synonyms per entity (having at least one synonym) in the NILK-sample dataset is also lower, compared to the other datasets.
	
	\section{Evaluation}
	\subsection{Comparison Methods}
	
	The main baseline methods for out-of-KB mention discovery are: (i) \textbf{Rule-based approach} (or Sieve-based): using carefully crafted rules (each as a ``sieve'') designed for biomedical texts, where NIL is detected when no in-KB entity can be linked to \cite{dsouza-ng-2015-sieve-based}. (ii) \textbf{Threshold-based approach}: setting a threshold on the EL system's prediction score for each candidate, if the score for the closest in-KB candidate is still below the threshold, the mention is out-of-KB \cite{ji2011TAC,chen2021lightweight}; this is applied with BM25 for candidate generation and BERT-based cross-encoder for candidate ranking and synonym enhancement (``$Th$-based BM25+BERT+syn'') \cite{ji2020bert}, where we tuned the threshold (as 0.85) and the domain-specific setting, as SapBERT \cite{liu2021sap} (``$Th$-based BM25+SapBERT+syn''), based on the validation set. (iii) \textbf{Feature-based NIL entity classification } (``Ft-based Classifer''): including string-matching features \cite{poibeau_entity_2013,dredze2010tac,mcnamee2009hltcoe}, entity contextual feature space, and embedding-based feature space \cite{wu2016features}; we also provide a feature-based baseline that uses the entity candidates and dynamic features (i.e., $v^{(cross)}_{m,e}$) in BLINKout (``Ft-based BLINKout $L^{(Ft)}_{NIL}$'').
	
	For model comparison, the default number of top-$k$ was set as 10 for the baselines using BM25, BLINK, and BLINKout, following \citet{ji2020bert}. In the ``$k$ and NIL rep tuned'' setting, both $k$ and NIL entity representation were tuned based on the validation set to optimise $F_{1_o}$ (see Section \ref{sensitive-settings} and Appendix \ref{param-top-k} for details in parameter tuning, with the optimal settings in Appendix \ref{exp-details-params}). 
	
	\textbf{Adapting BLINK without Training}. We also implemented both threshold-based and the NIL entity representation approach to a BLINK model trained from all in-KB entities (``$Th$-based BLINK''\footnote{The out-of-KB threshold is placed at the cross-encoder. We also ensure that the NIL entity is in the bi-encoder's top-$k$ candidates by replacing the $k$-th candidate to NIL, thus adding an extra threshold for the bi-encoder does not affect the results.} and ``NIL-rep-based BLINK''), either or not using synonyms (``+syn'').
	
	\subsection{Evaluation Metrics}
	In the conventional EL setting, which does not discriminate between in-KB and out-of-KB mentions, the \textit{overall} (or in-KB$+$out-of-KB) accuracy ($A$), precision ($P$), recall ($R$), and $F_1$ scores are the same, i.e., $A = P = R = F_1 = \frac{TP}{|M|}$, where $TP$ and $|M|$ denote the number of correctly linked mentions and all the target mentions, resp. \cite{shen2014entity,sevgili_neural_2022}.
	
	In this work, we propose to use \textit{out-of-KB} precision ($P_o$), recall ($R_o$), and $F_1$ ($F_{1_{o}}$) scores to measure how well out-of-KB mentions are detected.
	Analogously, we can calculate the \textit{in-KB} precision ($P_{in}$), recall ($R_{in}$), and $F_{1_{in}}$ scores.
	\begin{equation}\label{eval-out-metrics}
		P_o = \frac{TP_o}{TP_o+FP_o}; R_{o} = \frac{TP_o}{TP_o+FN_o}
	\end{equation}
	
	\begin{equation}
		P_{in} = \frac{TP_{in}}{TP_{in}+FP_{in}}; R_{in} = \frac{TP_{in}}{TP_{in}+FN_{in}}
	\end{equation}
	where $TP_o$, $FP_o$, and $FN_o$ (resp., $TP_{in}$, $FP_{in}$, and $FN_{in}$)  are the numbers of true positive, false positive, and false negative out-of-KB (resp., in-KB) mentions. $TP_{in}$ refers to the number of mentions that are predicted with the correct entities in the KB instead of mentions that are simply predicted as in-KB.
	$F_1$ is the harmonic mean of $P$ and $R$.
	
	\subsection{Results}
	\begin{table*}[t!]
		\small
		\center
		\renewcommand{\arraystretch}{0.79}
		\setlength{\tabcolsep}{2.9pt}
		\begin{tabular}{l||l|lll||l|lll||l|lll||l|lll||l|lll}
			\toprule
			& \multicolumn{4}{c||}{ShARe/CLEF 2013}                    & \multicolumn{4}{c||}{MM-pruned-0.1} & \multicolumn{4}{c||}{MM-pruned-0.2} & \multicolumn{4}{c||}{MM-2014AB} & \multicolumn{4}{c}{NILK-sample}\\
			\midrule
			& $A$ & $P_{o}$ & $R_{o}$ & $F_{1_{o}}$ & $A$ & $P_{o}$ & $R_{o}$ & $F_{1_{o}}$ & $A$ & $P_{o}$ & $R_{o}$ & $F_{1_{o}}$ & $A$ & $P_{o}$ & $R_{o}$ & $F_{1_{o}}$ & $A$ & $P_{o}$ & $R_{o}$ & $F_{1_{o}}$ \\
			Rule-based \scriptsize{(Sieve-based)}                         & 90.0 & 84.7               & 92.6           & 88.5        & \textbf{83.8} & 36.3 & \textbf{94.8} & 52.5 & \textbf{85.3} & 63.4 & \textbf{97.1} & 76.7  & \textbf{82.9} & 26.6 & \textbf{98.1} & 41.9 & 72.1 & 6.6 & \textbf{86.7} & 12.2 \\
			Ft-based \scriptsize{Classifier}            & 36.0 & 49.1            & 71.6 & 58.2 & 19.1 & 11.2 & 69.7 & 19.2 & 25.8 & 27.5            & 63.9 & 38.4 & 20.1 & 7.2 & 37.9 & 12.1 & 24.2 & 1.6 & 70.3 & 3.0 \\
			Ft-based \scriptsize{BLINKout $L^{(Ft)}_{NIL}$}         & 60.9  & 61.4            & \textbf{95.0} & 74.6 & 42.3 &  14.0 & 92.3 & 24.2  & 48.0 & 39.3 & 95.9 & 55.7 & 43.7 & 8.9 & 85.4 & 16.1 & 45.2 & 1.8 & 61.4 & 3.5\\
			$Th$-based \scriptsize{BM25+BERT+syn} & 69.5 & 64.9               & 55.8            & 60.0        & 72.3 & 22.8 & 64.5 & 33.7  & 70.2 & 46.2 & 60.7 & 52.5 & 70.6 & 9.7 & 26.2 & 14.1 & 77.4 & 7.0 & 39.2 & 11.9 \\
			$Th$-based \scriptsize{BM25+SapBERT+syn} & 76.2 & 71.3               & 64.7            & 67.8        & 75.6 & 28.6 & 84.7 & 42.2  & 74.7 & 51.3 & 72.5 & 60.1 & 71.4 & 16.9 & 81.6 & 28.0 & 75.8 & 7.6 & 70.3 & 13.8 \\
			\midrule
			BLINK (BERT) & 53.0 & - & - & - & 67.0 & - & - & - & 57.2 & - & - & - & 62.4 & - & - & - & 88.0 & - & - & - \\
			\hspace{2mm} + syn & 58.1 & - & - & - & 70.4 & - & - & - & 60.2 & - & - & - & 68.8 & - & - & - & 87.8 & - & - & - \\
			$Th$-based BLINK                                      & 72.4 & 65.4               & 67.7            & 66.5  & 68.1 & 21.9 & 42.6 & 28.9 & 71.1 & 51.6               & 77.6            & 62.0 & 65.7 & 21.7 & 74.8 & 33.6 & 88.1 & 17.8 & 49.4 & 26.2  \\
			\hspace{2mm} + syn      & 80.1                                & 74.4               & 76.2            & 75.3 & 72.7 & 25.5 & 54.8 & 34.8   & 69.5                                & 47.7               & 60.5            & 53.4 & 71.1 & 22.2 & 48.5 & 30.5 & 87.8 & 16.3 & 44.3 & 23.8  \\
			NIL-rep-based BLINK                      & 74.8  & 67.6               & 89.2            & 76.9 & 69.7 & 24.8 & 47.1 & 32.5  & 71.9  & 56.9               & 74.9            & 64.7 & 65.7 & 19.1 & 80.6 & 30.9 & 87.9 & 16.3 & 24.7 & 19.6 \\
			\hspace{2mm} + syn                        & 80.3 & 79.8               & 73.5            & 76.5 & 72.6 & 27.8 & 40.7 & 33.0  & 73.8 & 54.4               & 77.1           & 61.7  & 71.4 & 18.6 & 74.8 & 29.8 &  87.9 & 16.2 & 55.7 & 25.0  \\
			\midrule
			BLINKout (- syn, BERT)                             & 77.7 & 74.1               & 91.2            & 81.8 & 62.3 & 18.4 & 62.6 & 28.5 & 68.8 & 51.1               & 80.7            & 62.6 & 68.2 & 25.2 & 78.6 & 38.1 & 88.1 & 24.3 & 49.4 & 32.6 \\
			BLINKout (BERT)     & 86.0                  & 85.3               & 87.9            & 86.6 & 70.5 & 33.1 & 58.7 &  42.3  & 75.2                  & 55.3               & 77.8            & 64.7  & 69.9 & 33.7 & 84.5 & 48.2 & 87.9 & 21.4 & 72.2 & 33.0 \\
			BLINKout (SapBERT)           & 89.9      & 90.0               & 89.8            & 89.9 & 81.8 & 53.5 & 54.8 & 54.1 & 81.4      & \textbf{72.6}               & 81.0            & 76.5  & 81.5 & 57.3 & 79.6 &    66.7 & 86.6 & 22.3 & 69.0 & 33.7 \\
			\hspace{2mm}+ $k$ and NIL rep tuned$^{*}$            & 90.8    & \textbf{91.4}               & 90.3            & \textbf{90.9}  & 83.3 & \textbf{58.4} & 65.2 & \textbf{61.6} & 84.8    & \textbf{72.6}               & 88.7            & \textbf{79.8}  & \textbf{82.9} & \textbf{86.2} & 72.8 & \textbf{79.0} & \textbf{90.7} & 33.0 & 70.3 & 44.9 \\ 
			\hspace{2mm}+ $L_{NIL}$   & \textbf{91.2} & 88.5                   & 92.5                & 90.5   & 83.1 & 47.0 & 75.5 & 57.9 & 83.0 & 65.2                   & 91.6               & 76.2 & 82.5 & 59.3 & 83.5 & 69.4 & 90.4 & \textbf{37.1} & 67.1 & \textbf{47.8} \\
			\bottomrule
		\end{tabular}
		\caption{Comparison results for Entity Linking  with \textit{out-of-KB} mentions (\underline{P}recision, \underline{R}ecall, $F_1$ scores w.r.t. out-of-KB mentions and  \textit{overall} \underline{A}ccuracy w.r.t. both in- and out-of-KB mentions). *The parameters $k$ and NIL rep are tuned using BLINKout (SapBERT) for ShARe/CLEF 2013 and MM datasets and BLINKout (BERT) for NILK-sample.}\label{out-of-KB-results}
		\vspace{-3mm}
	\end{table*}
	
	\textbf{Out-of-KB Metrics}. Table \ref{out-of-KB-results} shows the main results for out-of-KB mention discovery. BLINKout, with SapBERT and for NILK-sample with BERT, achieved the top out-of-KB $F_1$ ($F_{1_o}$) on all datasets. 
	
	The rule-based method (Sieve-based) \cite{dsouza-ng-2015-sieve-based} was effective on biomedical datasets, given the regular syntactic structure of the biomedical texts and concepts, but less effective on the general domain (NILK-sample). The rule-based approach resulted in high out-of-KB recall scores, but still lacked in out-of-KB precision and $F_{1_o}$ compared to BLINKout; this is because the former tends to result in a large number of out-of-KB mentions using the ``NIL as no candidate'' rule. 
	
	Feature-based methods, facing the challenge of imbalanced binary classification, also tend to predict a larger number of NILs and result in a high out-of-KB recall, but low precision and $F_1$ scores. 
	
	Threshold-based approaches provide higher out-of-KB precision than feature-based, but lower metrics than the rule-based approach. The setting using synonyms with bi-encoder (``BLINK+syn'') instead of BM25 (``BM25+BERT+syn'') greatly improved the in-KB results and thus the out-of-KB results based on a threshold.
	
	NIL-rep-based approaches with fixed NIL entity representations performed better than or on par with threshold-based approaches. The special token \texttt{[NIL]}, even fixed (unsupervised), can discriminate NIL from in-KB entities using the weights in pre-trained LM.
	
	BLINKout outperformed all the baselines on all the datasets. A large margin was achieved compared to the rule-based (second best for biomedical datasets, i.e., except for NILK-sample): for MM-2014AB and NILK-sample, by about 30-40\% improvement of $F_{1_o}$.
	
	For NILK-sample, the improvement with synonyms was less obvious. This is due to the zero-shot nature of the dataset and inadequacy of synonyms: first, NILK is a zero-shot entity linking dataset, while synonyms most benefit seen entities rather than unseen entities\footnote{We also tested the zero-shot (ZS) version of the other four datasets (by removing testing mentions linked to overlapped entities to the training set), and found that using synonyms may not improve much (compared to the original data setting) or may even impede the performance of linking, see results in Appendix \ref{appendix:zs-EL}.}; second, the percentage of entities having a synonym is much lower in NILK-sample (25\%) than the other datasets (99\%) and the average number of synonyms per entity is lower.
	
	\textbf{Ablation Study}. The contributing components in BLINKout are: (i) \textbf{fine-tuned NIL entity representation} with NIL-labelled training data, e.g., with 3-18\% improvement of $F_{1_o}$ between ``BLINKout (BERT)'' and ``NIL-rep-based BLINK+syn''; (ii) \textbf{synonym enhancement}, e.g., with 2-14\% improvement of $F_{1_o}$ between ``BLINKout (BERT)'' and ``BLINKout (-syn, BERT)'', except for a marginal improvement (of 0.4\%) on the NILK-sample dataset; (iii) \textbf{domain-specific LMs}, e.g., with 3-19\% improvement of $F_{1_o}$ between ``BLINKout (SapBERT)'' and ``BLINKout (BERT)'', except for a marginal improvement (of 0.7\%) on the NILK-sample dataset. 
	
	\textbf{Overall Accuracy}. For the overall accuracy on both in-KB and out-of-KB mentions ($A$), the proposed models perform the best or competitively in all datasets. This shows that the approaches to discover out-of-KB mentions do not compromise the performance of in-KB entities. In-KB EL results are in Table \ref{in-KB-results} in Appendix \ref{appendix:in-KB}.
	
	\subsection{Sensitive Settings}\label{sensitive-settings}
	
	The results of BLINkout are sensitive to the settings below.
	
	\textbf{The Number of Top-\textit{k} Candidates}. Out-of-KB metric scores generally have a positive correlation with $k$ when it is below a limit. A higher top-$k$ may make it harder to train a model to discriminate out-of-KB from the remaining classes using the cross-encoder; this especially decreases the performance when there are limited out-of-KB samples for training (e.g., in MM-2014AB). A lower top-$k$ is likely to decrease in-KB EL results. 
	Details are in Appendix \ref{param-top-k}.
	
	\textbf{NIL Entity Representation}. NIL representation sets a prior anchor point in the embedding space for the bi-encoder and is used in the cross-encoder for candidate ranking. Results on ShARe/CLEF 2013 and NILK-sample are displayed in Table \ref{NIL-rep-results}: \texttt{[NIL]} representation performed the best in most settings, either fixed or fine-tuned.
	
	\begin{table}[t!]
		\small
		\center
		\renewcommand{\arraystretch}{0.79} 
		\setlength{\tabcolsep}{2.8pt}
		\begin{tabular}{llll|lll}
			\toprule
			& \multicolumn{3}{c}{ShARe/CLEF 2013} & \multicolumn{3}{c}{NILK-sample}        \\
			\cline{2-7}
			& A & $F_{1_{o}}$ & $F_{1_{in}}$ & A & $F_{1_{o}}$ & $F_{1_{in}}$  \\
			\midrule
			\textbf{Fixed rep:} \\
			\multicolumn{2}{l}{\textbf{NIL-rep-based BLINK+syn}}\\
			\hspace{2mm}NIL & 70.0               & 52.3 & 75.4 & 87.3 & 3.3 & 88.0                \\
			\hspace{2mm}NIL {[}ENT{]} It is a NIL option.  & 62.7               & 25.6 & 71.5 & 87.3 & 1.2 & 88.0\\
			\hspace{2mm}{[}NIL{]} & \textbf{80.3} & \textbf{76.5} & \textbf{82.0} & \textbf{87.9} & \textbf{25.0} & \textbf{90.1} \\
			\hspace{2mm}{[}NIL{]} {[}ENT{]} It is a NIL option. & 74.5 & 64.9 & 78.0 & 87.8 & 24.7 & 89.4\\
			\hspace{2mm}{[}NIL{]} {[}ENT{]} It is a {[}NIL{]} option. & 76.5 & 69.2 & 79.4 & 87.8 & 24.1 & 89.7 \\
			\midrule\midrule
			\textbf{Fine-tuned rep:} \\
			\textbf{BLINKout (BERT), $k$ tuned} \\
			\hspace{2mm}{[}NIL{]} & \textbf{88.7} & \textbf{89.1} & \textbf{88.5} & \textbf{90.7} & 44.9 & \textbf{91.8} \\
			\hspace{2mm}{[}NIL{]} {[}ENT{]} It is a {[}NIL{]} option. & 87.6 & 88.2 & 87.2 & 90.3 & \textbf{46.2} & 91.1 \\
			\bottomrule
		\end{tabular}
		\caption{Comparison results among NIL entity representations (``rep''), either fixed or fine-tuned}\label{NIL-rep-results}
		\vspace{-3mm}
	\end{table}
	
	\textbf{Domain-Specific LMs}. In-domain and knowledge-enhanced LMs, SapBERT (and also PubMedBERT \cite{gu2022pmbert}), obtained better in- and out-of- KB results for datasets in the biomedical domain.
	
	\section{Qualitative Analysis}\label{qualitative-analysis}
	
	\begin{table*}[t!]
		\footnotesize
		\renewcommand{\arraystretch}{0.82} 
		\setlength{\tabcolsep}{2.4pt}
		\begin{tabular}{p{1.6cm}p{5cm}p{1.2cm}p{2.2cm}p{2.2cm}p{2.2cm}p{2.2cm}}
			\toprule
			Data w/ row ID (starts from 0)                               & \textbf{Mention} in context                                                                                                                                                                                                                                                                                                   & Gold & $Th$-based BLINK+syn                                                                                                                              & NIL-rep-based BLINK+syn                                                     & BLINKout (tuned)                                                                           & BLINKout (tuned)+$L_{NIL}$                                                   \\
			\midrule
			ShARe/ CLEF 2013 test-2353 & Discharge Diagnosis: \textbf{PEA arrest} of unclear etiology   & NIL                                                                                                                                                                                                                                                                & \textcolor{red}{ Electromechanical dissociation (C0340861) 1.000\textgreater $Th_{cross}$ (0.95)}                                                  & \textcolor{red}{ Electromechanical dissociation (C0340861) 1.000} (where NIL 0.000, 2nd) & \textcolor{red}{ Electromechanical dissociation (C0340861) 0.888} (where NIL 0.001, 2nd) & NIL 0.718                                                                                                        \\
			ShARe/ CLEF 2013 test-5083 & Low dose beta blockade \textbf{decreased HR} to 95 in sinus and   synthroid had been restarted.  & Bradycardia (C0428977)                                                                                                                                                                                                                              & \textcolor{red}{ NIL (Cardiac murmur, intensity   grade I/VI (C0232249) 0.191\textless $Th_{cross}$ (0.95))
			} (Gold not in top-$k$)                   & \textcolor{red}{ NIL 0.973} (Gold not in top-$k$)                    & Bradycardia (C0428977) 0.989                                                               & \textcolor{red}{ NIL 0.947} (where Bradycardia   (C0428977) 0.053, 2nd)          \\
			\midrule
			MM-pruned-0.1 test-14            & Dietary antioxidants may play an   important role in the prevention of  \textbf{bone loss} and associated fractures   by reducing levels of oxidative stress.        & NIL                                                                                                                                                                & NIL (Tooth   eruption disorder (C0012767)  0.627\textless $Th_{cross}$ (0.80))                                                         & \textcolor{red}{ Tooth eruption disorder   (C0012767) 0.426} (where NIL 0.320, 2nd)        & NIL 1.000                                                                                  & NIL 0.997                                                                                                  \\
			MM-pruned-0.1 test-35            & ... including weight and fat   gain, glucose intolerance, \textbf{hypertriglyceridemia}, abnormal adipocytokine levels, hypertension, and adiponectin   and leptin gene expression and epigenetic changes.        & NIL                                                                                                               & \textcolor{red}{ Hyperglycemia (C0020456) 1.000\textgreater $Th_{cross}$ (0.80)}                                                                                        & \textcolor{red}{ Hyperglycemia (C0020456) 1.000} (where NIL 0.000, 3rd)                    & \textcolor{red}{ Dyslipidemias (C0242339) 0.564} (where NIL 0.011, 3rd)                  & \textcolor{red}{ Dyslipidemias (C0242339) 0.553} (where NIL 0.205, 3rd)   \\     
			\midrule
			NILK-sample test-68 & The ancient cities of Friesland are shown below:    Dutch  West Frisian  Charter granted    ...    Workum   Warkum  1399    \textbf{Bolsward}    Boalsert  1455    Harlingen   Harns  1234    Franeker   Frjentsjer  1374 & NIL & NIL (Ferdinand Bol (Q374039) 0.403<$Th_{cross}$ (0.45)) & \textcolor{red}{Ferdinand Bol (Q374039) 0.480} (where NIL 0.276, 2nd) & NIL 0.377 & NIL 0.191 \\
			NILK-sample test-2054 & ... \textbf{70720 Davidskillman}        Named in honor of David R. Skillman (b. 1945) for his decades-long contributions to asteroid searching, stellar binary star systems and as lead systems engineer for the Hubble Space Telescope at Goddard Space Flight Center. & NIL & \textcolor{red}{DAVID (Q5204342) 0.980>$Th_{cross}$ (0.45)} & \textcolor{red}{DAVID (Q5204342) 0.987} (where NIL 0.019, 2nd) & NIL 0.798 & NIL 0.965 \\
			NILK-sample test-755 & TMS Entertainment (known for ''Lupin III'' and ''Detective Conan'') would animate the popular animated series ''Tiny Toon Adventures'' and ''\textbf{Rainbow Brite}''. ... & NIL & \textcolor{red}{Rainbow Brite and the Star Stealer (Q2625526) 1.000>$Th_{cross}$ (0.45)} & \textcolor{red}{Rainbow Brite and the Star Stealer (Q2625526) 1.000} (where NIL 0.000, 2nd) & \textcolor{red}{Rainbow Brite and the Star Stealer (Q2625526) 1.000} (where NIL 0.000, 2nd) & \textcolor{red}{Rainbow Brite and the Star Stealer (Q2625526) 0.999} (where NIL 0.001, 2nd) \\
			\bottomrule          
		\end{tabular}
		\caption{Examples of out-of-KB mention discovery from clinical notes (ShARe/CLEF 2013), biomedical publications (MedMentions), and Wikipedia articles (NILK) with the top prediction from four BERT-based Entity Linking models: Threshold-based BLINK, NIL representation based BLINK, BLINKout (tuned), and BLINKout (tuned) with joint learning ($L_{NIL}$). Normalised prediction score (after softmax) and the rank (if not 1st) are displayed after the predicted entity. Wrong predictions are marked with \textcolor{red}{red}. For the three datasets, ShARe/CLEF 2013, MM-pruned-0.1, NILK-sample, the threshold $Th_{cross}$ for $Th$-based BLINK was 0.95, 0.80, 0.45, and the tuned number of top-$k$ for BLINKout was 150, 50, 50, resp. }\label{example-preds}
	\end{table*}
	
	We selected samples from the test set in the datasets regarding erroneous predictions for out-of-KB mentions, displayed in Table \ref{example-preds}. NILK-sample dataset has many text sequences rendered from Tables (see rows 5-6), which provides a different context compared to other datasets, and fine-grained entities such as names of cities in other languages, asteroids, and animated series (see rows 5-7).
	
	BLINKout models can correctly identify NIL mentions, while false but similar in-KB entities were predicted in $Th$- and/or NIL-rep- based methods (see rows 3, 5-6). BLINKout with joint learning affects the prediction score for NIL, which can result in both true or false positive NIL predictions (\textit{cf.} row 1 and 2). The NIL-rep-based approach can sensitively identify NIL mentions with a high rank (if not predicted as the top, in rows 1, 3-7), without training with NIL-labelled mentions. The threshold-based approach may likely result in false negative predictions for out-of-KB mentions when the incorrect but similar entity was predicted with a very high score near 100\% (as in rows 1, 4, 6-7). Finally, there are challenging out-of-KB entities which are very similar to in-KB entities in the KB pruning based dataset, MM-pruned-0.1, and fine-grained entities in NILK-sample, so that all models yielded a wrong prediction, but with a high rank of NIL as the 2nd or 3rd (in rows 4, 7).
	
	\section{Discussion}
	\textbf{Results w.r.t. Dataset Construction Strategies}. The out-of-KB performance of BLINKout was the highest on the Manual Labelling data, ShARe/CLEF 2013, and lowest on the KB Versioning data, NILK-sample. This is partially related to the percentage of out-of-KB mentions: the lower the percentage in the training data, the more challenging their detection. On the more realistic KB Versioning dataset, BLINKout outperforms the rule-based approach with a larger gap compared to datasets created with the other strategies. 
	
	\textbf{NIL Entity Representation}.
	The results show that NIL entity representation based methods perform better or on par with the threshold-based method, and also have a high rank of NIL among in-KB entities for out-of-KB mentions. BLINKout with the fine-tuned \texttt{[NIL]} representation yielded a large margin of improvement. It is safer to use non-natural language tokens, i.e., \texttt{[NIL]}, rather than ``NIL'' to avoid confusion with other vocabularies.
	
	\textbf{Joint Learning}.
	Using $L_{NIL}$ allows to penalise the situation that a NIL mention being wrongly classified as an in-KB entity, or vice-versa. This may increase the out-of-KB recall $R_{o}$, but harm the precision $P_o$ and $F_{1_o}$ scores (except for NILK-sample), as shown in Table \ref{out-of-KB-results}. Joint learning obtained the best overall accuracy on ShARe/CLEF 2013 and best out-of-KB $F_{1}$ for NILK-sample. More effective use of $L_{NIL}$ warrants further studies. 
	
	\textbf{Synonyms Enhancement}. Synonyms help address the entity variant problem and further differentiate in-KB entities from NIL after the fine-tuning. Our synonym enhancement on the bi-encoder and cross-encoder greatly improved the out-of-KB $F_1$ scores across most datasets over BLINK. In the zero-shot EL scenario (e.g., NILK-sample data and zero-shot test sets of other data in Appendix \ref{appendix:zs-EL}), synonyms may be less useful given that synonyms of unseen entities are not presented during training. Making use of synonyms in the ZS setting warrants further studies.
	
	\section{Related Work}
	
	\textbf{Entity Linking (EL) for KB Construction}. EL has been extensively studied, see reviews in \citet{shen2014entity,shen2021eldeep,sevgili_neural_2022}. The work in \citet{poibeau_entity_2013} focuses on a KB-centric view of EL, i.e., 
	EL as a key step in KB construction and maintenance. Most relevant to KB maintenance is the identification of out-of-KB or NIL mentions so that they can be placed into the KB \cite{poibeau_entity_2013}. Many recent studies in EL only considered in-KB entities \cite{bhowmik2021,wu2020blink,basaldella2020cometa}. Out-of-KB mention discovery is also distinct from zero-shot EL \cite{wu2020blink,Ristoski2021kgzselEbay}, as the latter targets in-KB entities.
	
	\textbf{Out-of-KB Mention and Entity Discovery}. Most studies apply the traditional threshold-based approach \cite{bunescu-pasca-2006,ji2020bert,chen2021lightweight,xu2022simple} or NIL classification with features \cite{wu2016features,poibeau_entity_2013,Zhang2020} and neural networks \cite{martins-etal-2019-joint,gu2021read}. We proposed NIL entity representation \& classification that utilizes the embedding representation in LMs, with BERT-based EL \cite{wu2020blink}.
	
	Recent studies form NIL clusters for the resolution of out-of-KB mentions into potential concepts or entities \cite{kassner2022edin,agarwal2021entity,heist2023nastylinker}. The NIL clusters are formed either with a threshold-based approach or as a post-processing of the NIL mentions after their discovery, as summarized in \citet{ji2011TAC}. Our methods on NIL entity representation \& classification are independent of clustering methods, and we leave the combination and comparison with them for a future study.
	
	\textbf{Out-of-KB EL Benchmarking}. 
	We summarized most of the NIL-labeled EL datasets in the introduction, including ShARe/CLEF 2013 \cite{suominen2013shareclef}, NILK \cite{Iurshina2022nilk}, NEEL 2015-2016 challenges \cite{rizzo2017NEEL}, and CLEF HIPE 2020 \cite{ehrmann2020extended}. The earliest dataset is TAC 2011 Knowledge Base Population Track \cite{ji2011TAC}, which has NIL mentions to enrich a KB derived from Wikipedia Infoboxes, however, is not freely available and the data size is small (around 1000 mentions in training and testing resp.). Another recent clinical note dataset is 2019-n2c2-MCN \cite{Luo2020}, which is not yet openly available for new users outside of the previous registration for the challenge. Another recent general domain dataset is EDIN-benchmark \cite{kassner2022edin} to enrich older versions of Wikipedia using text mentions from news articles. EDIN-benchmark requires an adaptation set of documents, which may not be always available. We summarized and applied strategies (manual labelling, KB pruning, KB versioning) for NIL-enhanced EL data creation. As far as we know, all previous studies have considered only one of the strategies and we are the first study encompassing all three strategies for dataset construction and benchmarking.
	
	Regarding evaluation, overall accuracy \cite{ji2020bert,chen2021lightweight,gu2021read} is commonly used, which does not reflect the full picture. As far as we know, our study is the first to apply \textit{out-of-KB} Precision, Recall, and $F_1$ scores.
	
	The research endeavour that mostly resembles ours is the ongoing work of \citet{moller2022knowledge} on EL for KB enrichment. This study targets news events to enrich Wikipedia. Similar to the plans in \citet{moller2022knowledge}, our future work will canonicalise the new mentions by, for example, grouping and naming, and placing them in the KB.
	
	\section{Conclusion}
	
	We introduced the task of out-of-KB mention discovery from texts and proposed BLINKout, which utilizes a dynamic NIL representation \& classification approach, enhanced with synonyms, founded on BERT-based Entity Linking. We also provided strategies, KB Pruning and KB Versioning, to construct out-of-KB datasets. The approach has been tested on datasets with various domains to enrich medical ontologies and WikiData. The method in this work, while extending the BERT-based Entity Linking (BLINK) approach \cite{wu2020blink}, also has the potential to be applied to recent end-to-end Entity Linking methods \cite{ayoola-etal-2022-refined}. 
	Future studies will focus on the canonicalisation and placement of out-of-KB entities in a KB.
	
	\noindent\textbf{Acknowledgements}. This work is supported by EPSRC projects, including ConCur (EP/V050869/1), OASIS (EP/S032347/1), UK FIRES (EP/S019111/1); and Samsung Research UK (SRUK).
	
	\appendix
	
	\section{Experimental Details }\label{exp-details-params}
	BLINKout was implemented using PyTorch Huggingface Transformers\footnote{\url{https://huggingface.co/docs/transformers/index}}, based on the original BLINK library\footnote{\url{https://github.com/facebookresearch/BLINK}}. The BM25+BERT baseline further used the Python library rank\_bm25\footnote{\url{https://pypi.org/project/rank-bm25/}}. Results on the Sieve-based approach were reproduced using the original JAVA-based implementation\footnote{\url{https://github.com/jennydsuza9/disorder-normalizer}}. All neural network models were trained using an NVIDIA Quadro RTX 8000 GPU card (48GB GPU memory). 
	
	\begin{table*}[t!]
		\small
		\center
		\renewcommand{\arraystretch}{0.79} 
		\setlength{\tabcolsep}{6.5pt}
		\begin{tabular}{llll|lll|lll|lll|lll}
			\toprule
			& \multicolumn{3}{c|}{ShARe/CLEF 2013}                    & \multicolumn{3}{c|}{MM-pruned-0.1} & \multicolumn{3}{c|}{MM-pruned-0.2} & \multicolumn{3}{c|}{MM-2014AB} & \multicolumn{3}{c}{NILK-sample} \\
			\midrule
			& $P_{in}$ & $R_{in}$ & $F_{1_{in}}$ & $P_{in}$ & $R_{in}$ & $F_{1_{in}}$ & $P_{in}$ & $R_{in}$ & $F_{1_{in}}$ & $P_{in}$ & $R_{in}$ & $F_{1_{in}}$ & $P_{in}$ & $R_{in}$ & $F_{1_{in}}$\\
			Rule-based \scriptsize{(Sieve-based)}                         & \textbf{93.1}               & 88.8           & 90.9        & \textbf{95.9} & 82.9 & \textbf{88.9} & \textbf{95.5} & 82.2 & \textbf{88.4} & \textbf{96.1} & 82.1 & \textbf{88.6} & 88.2 & 71.9 & 79.2 \\
			Ft-based \scriptsize{Classifier}            & 19.0            & 24.4 & 21.4 & 26.6 & 14.9 & 19.0 & 24.1 & 15.8 & 19.1 & 25.0 & 19.2 & 21.7 & 70.9 & 23.5 & 35.3\\
			Ft-based \scriptsize{BLINKout $L^{(Ft)}_{NIL}$}         & 60.4           & 44.6 & 51.3 &  72.1 & 38.1 & 49.9  & 57.0 & 35.5 & 43.7 & 78.1 & 41.4 & 54.1 & 88.6 & 45.0 & 59.7 \\
			$Th$-based \scriptsize{BM25+BERT+syn} & 71.3               & 76.3            & 73.6        & 86.2 & 73.0 & 79.0 & 79.2 & 72.6 & 75.8 & 80.5 & 73.0 & 76.6 & 83.8 & 78.0 & 80.8 \\
			$Th$-based \scriptsize{BM25+SapBERT+syn} & 78.2               & 81.7            & 79.9        & 88.7 & 75.1 & 81.3  & 84.4 & 75.3 & 79.6 & 89.7 & 70.9 & 79.2 & 86.6 & 75.9 & 80.9 \\
			\midrule
			BLINK (BERT) & 53.0 & 78.4 & 63.2 & 67.0 & 72.6 & 69.7 & 57.2 & 72.2 & 63.8 & 62.4 & 65.8 & 64.1 & 88.0 & 89.3 & 88.6\\
			\hspace{2mm}+ syn & 58.1 & 85.8 & 69.3 & 70.4 & 76.3 & 73.2 & 60.2 & 76.0 & 67.2 & 68.8 & 72.6 & 70.6 & 87.8 & 89.1 & 88.4\\
			$Th$-based BLINK                                      & 76.0               & 74.7            & 75.3  & 76.3 & 70.2 & 73.2 & 79.9 & 69.3 & 74.2 & 75.3 & 65.3 & 69.9 & 91.1 & 88.7 & 89.9   \\
			\hspace{2mm}+ syn                             & 82.9               & 81.9            & 82.4 & 82.2 & 74.2 & 78.0  & 77.2 & 71.8 & 74.4 & 77.3 & 72.3 & 74.7 & 90.8 & 88.5 & 89.6  \\
			NIL-rep-based BLINK                       & 80.2               & 68.0            & 73.6  & 77.4 & 71.6 & 74.4  & 77.5 & 71.0 & 74.1 & 78.8 & 64.9 & 71.2 & 87.9 & 89.6 & 88.9  \\
			\hspace{2mm}+ syn                      & 80.5               & 83.6            & 82.0 & 78.3 & 75.3 & 76.8 & 81.0 & 74.5 & 77.6 & 85.3 & 71.3 & 77.7 & 91.8 & 88.4 & 90.1 \\ 
			\midrule
			BLINKout (- syn, BERT)                             & 80.2               & 71.3            & 75.4 & 77.9 & 62.2 & 69.2 & 77.4 & 65.6 & 71.0 & 76.5 & 67.6 & 71.8 & 90.1 & 88.7 & 89.4    \\
			BLINKout (BERT)      & 86.4               & 85.2            & 85.8 & 76.5 & 71.5 &  73.9 & 83.4 & 74.5 & 78.7 & 75.3 & 69.1 & 72.1 & 91.5 & 88.2 & 89.8  \\
			BLINKout (SapBERT)$^*$           & 89.9               & 90.0            & 90.0 & 84.3 & \textbf{84.1} & 84.2 & 84.1 & 81.5 & 82.8 & 83.4 & 81.6 &    82.5 & 89.7 & 86.9 & 88.3  \\
			\hspace{2mm}+ $k$ and NIL rep tuned            & 90.5               & \textbf{91.0}            & 90.7  & 87.1 & 83.3 & 85.1 & 88.9 & \textbf{83.7} & 86.2 & 82.8 & \textbf{83.5} & 83.1 & \textbf{92.6} & \textbf{91.0} & \textbf{91.8} \\ 
			\hspace{2mm}+ $L_{NIL}$  & 92.5                & 90.5  & \textbf{91.5} & 88.2 & 83.7 & 85.9 & 90.3 & 80.7 & 85.2 & 84.3 & 82.4 & 83.4 & 91.9 & 90.8 & 91.3\\
			\bottomrule
		\end{tabular}
		\caption{Comparison results for \textit{in-KB} Entity Linking (\underline{P}recision, \underline{R}ecall, $F_1$ scores for in-KB entities). * The parameters $k$ and NIL rep are tuned using BLINKout (SapBERT) for ShARe/CLEF 2013 and MM datasets, and BLINKout (BERT) for NILK-sample.}\label{in-KB-results}
		\vspace{-3mm}
	\end{table*}
	
	We tuned the hyperparameters by optimising $F_{1_o}$ on the validation set. We generally tuned the best $k \in \{5,10,20,50,100,150,200\}$ for each dataset and selected the best NIL representation. The default $k$ for model comparison was set as 10 for Share/CLEF 2013 following \citet{ji2020bert} and MedMention datasets, and 4 for the NILK-sample dataset following \citet{Iurshina2022nilk}. The best BLINKout model had $k$ as 150 for ShARe/CLEF 2013, 50 for MM-pruned-0.1, MM-2014AB, and NILK-sample, and 100 for MM-pruned-0.2. The NIL representation was \texttt{[NIL]} for the datasets except \texttt{[NIL]} with \texttt{[NIL]}-definition for NILK-sample (only when using the fixed setting) and MM-pruned-0.1. For the threshold-based BLINK method, the threshold $Th_{cross}$ was 0.95, 0.55, 0.80, 0.95, 0.45 for the datasets ShARe/CLEF 2013, MM-2014AB, MM-pruned-0.1, MM-pruned-0.2, and NILK-sample, resp. 
	
	\begin{figure}[t!]
		\includegraphics[width=0.45\textwidth]{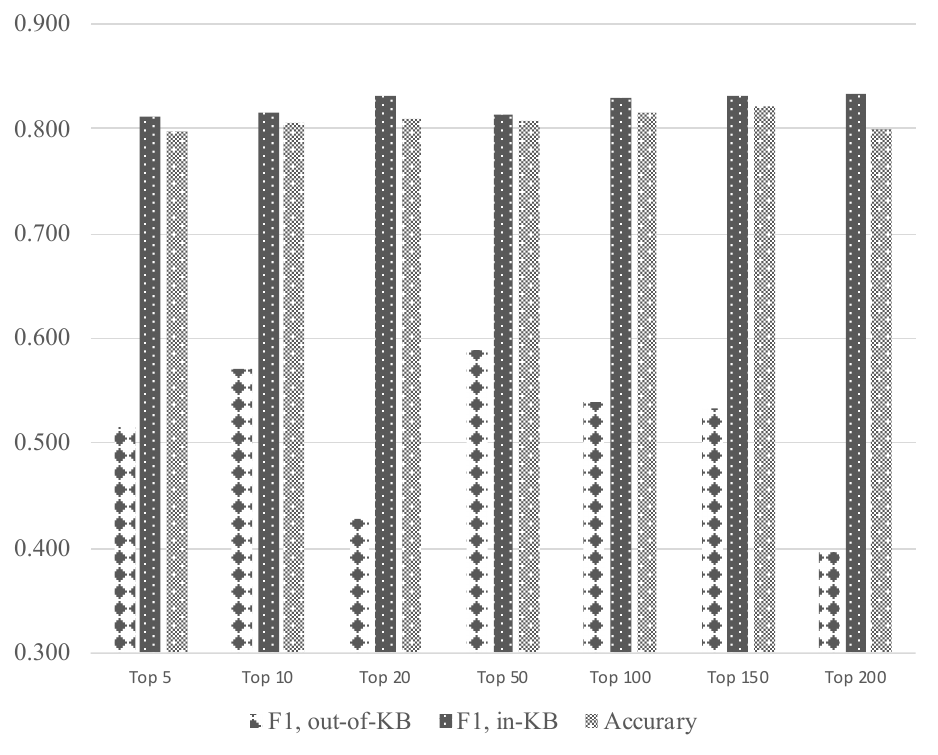}
		\caption{Out-of-KB $F_1$, in-KB $F_1$, and overall accuracy w.r.t. the number of top-$k$ candidates using BLINKout (SapBERT) on MM-2014AB validation set}\label{topk-cross-plot}
		\vspace{-3mm}
	\end{figure}
	
	We followed \citet{wu2020blink} to use a maximum of 32 tokens for a mention in context and 128 for an entity candidate (including its synonyms, 32 for NILK-sample), learning rate as 3e-05, bi-encoder training batch size as 16, cross-encoder training batch size as 1. We used the ``large'' version of LM (e.g., BERT-large, 680M parameters) in the bi-encoder and ``base'' version of LM (e.g., BERT-base, 110M parameters) in the cross-encoder. For SapBERT and PubMedBERT, we used ``base'' version for the bi-encoder as no ``large'' is available.
	
	The parameter in joint learning, $\lambda_{NIL}$, was tuned to 0.25 (as in \cite{gu2021read}) for ShARe/CLEF 2013 and MM-pruned-0.2 datasets, to 0.2 for MM-pruned-0.1, 0.05 for MM-2014AB, and 0.01 for NILK-sample, based on the validation set. We find that value of $\lambda_{NIL}$ is near to the percentage of out-of-KB mentions in Table \ref{data-stats} for each dataset.
	
	We optimised the bi-encoder and cross-encoder using AdamW \cite{loshchilov2018decoupled}, with 3 and 4 epochs, resp. (except for NILK-sample, 1 epoch each due to the large data size). We trained all models with fixed random seeds to obtain reproducible results and the actual variance was low with random seeds, less than 1\% (e.g., 0.85\% on MM-2014AB with BLINKout, $k$=10, over three runs).
	
	\textbf{Time Used in Training and Inference}. The training of bi-encoder with SapBERT on ShARe/CLEF 2013 took approx. 0.63 hour; cross-encoder with SapBERT took approx. 0.88 hour when $k$=10 and approx. 8.67 hours when $k$=150. The inferencing stage is very fast compared to the training: inferencing with a new mention took approx. 0.06 and 0.57 second for $k$=10 and $k$=150, resp.
	
	\section{Results w.r.t. Number of Top-$k$}\label{param-top-k}
	Figure \ref{topk-cross-plot} displays the change of $F_{1_o}$, $F_{in}$, and $A$ with respect to the number of $k$ on MM-2014AB with the BLINKout model (SapBERT).
	
	\begin{table}[t!]
		\small
		\center
		\renewcommand{\arraystretch}{0.79} 
		\setlength{\tabcolsep}{4.8pt} 
		\begin{tabular}{llll|lll}
			\toprule
			$k$=10 & \multicolumn{3}{c|}{ShARe/CLEF 2013-ZS} & \multicolumn{3}{c}{MM-pruned-0.1-ZS}                     \\
			\midrule
			& A & $F_{1_{o}}$ & $F_{1_{in}}$ & A & $F_{1_{o}}$ & $F_{1_{in}}$ \\
			BLINKout (-syn, BERT)                         & 78.0               & \textbf{90.7}            & 41.3  & \textbf{48.2} & 45.1 & \textbf{49.7}       \\
			BLINKout (BERT)                         & \textbf{80.3}               & 90.2            & \textbf{54.9}  & 43.8               & \textbf{51.6}            & 41.0      \\
			Diff in ZS setting & +2.3 & -0.5 & +13.6 & -4.4 & +6.5 & -8.7 \\
			Diff in original setting & +7.3 & +4.8 & +10.4 & +8.2 & +13.8 & +4.7\\
			\bottomrule
		\end{tabular}
		\caption{Comparison results on zero-shot (ZS) testing sets on BLINKout w.r.t. synonym enhancement. Their differences (``Diff'') in the original data setting is based on Table \ref{out-of-KB-results} and \ref{in-KB-results}.}\label{ZS-results} 
		\vspace{-3mm}
	\end{table}
	
	\section{Results for In-KB Entities}\label{appendix:in-KB}
	In-KB EL results on the four datasets are presented in Table \ref{in-KB-results}.
	
	\section{Zero-shot EL and Synonyms}\label{appendix:zs-EL} 
	Zero-shot entity linking results w.r.t. the use of synonyms to enhance BLINKout are presented in Table \ref{ZS-results}. Results show that in the zero-shot setting of the testing sets, i.e., after removing testing mentions having overlapped entities to the training set, the performance of out-of-KB mention discovery (e.g., $F_{1_o}$) with BLINKout is less sensitive to synonym enhancement. The inference of out-of-KB entities may be affected when ranking the unseen entities, which have abundant synonyms not presented during training.
	
	\bibliographystyle{ACM-Reference-Format}
	\bibliography{ent-link-kg}


\begin{thebibliography}{48}


\ifx \showCODEN    \undefined \def \showCODEN     #1{\unskip}     \fi
\ifx \showDOI      \undefined \def \showDOI       #1{#1}\fi
\ifx \showISBNx    \undefined \def \showISBNx     #1{\unskip}     \fi
\ifx \showISBNxiii \undefined \def \showISBNxiii  #1{\unskip}     \fi
\ifx \showISSN     \undefined \def \showISSN      #1{\unskip}     \fi
\ifx \showLCCN     \undefined \def \showLCCN      #1{\unskip}     \fi
\ifx \shownote     \undefined \def \shownote      #1{#1}          \fi
\ifx \showarticletitle \undefined \def \showarticletitle #1{#1}   \fi
\ifx \showURL      \undefined \def \showURL       {\relax}        \fi
\providecommand\bibfield[2]{#2}
\providecommand\bibinfo[2]{#2}
\providecommand\natexlab[1]{#1}
\providecommand\showeprint[2][]{arXiv:#2}

\bibitem[Agarwal et~al\mbox{.}(2021)]%
        {agarwal2021entity}
\bibfield{author}{\bibinfo{person}{Dhruv Agarwal}, \bibinfo{person}{Rico
  Angell}, \bibinfo{person}{Nicholas Monath}, {and} \bibinfo{person}{Andrew
  McCallum}.} \bibinfo{year}{2021}\natexlab{}.
\newblock \showarticletitle{Entity Linking and Discovery via Arborescence-based
  Supervised Clustering}.
\newblock \bibinfo{journal}{\emph{CoRR}}  \bibinfo{volume}{abs/2109.01242}
  (\bibinfo{year}{2021}), \bibinfo{pages}{1–12}.
\newblock
\showeprint[arXiv]{2109.01242}
\urldef\tempurl%
\url{https://arxiv.org/abs/2109.01242}
\showURL{%
\tempurl}


\bibitem[Ayoola et~al\mbox{.}(2022)]%
        {ayoola-etal-2022-refined}
\bibfield{author}{\bibinfo{person}{Tom Ayoola}, \bibinfo{person}{Shubhi Tyagi},
  \bibinfo{person}{Joseph Fisher}, \bibinfo{person}{Christos
  Christodoulopoulos}, {and} \bibinfo{person}{Andrea Pierleoni}.}
  \bibinfo{year}{2022}\natexlab{}.
\newblock \showarticletitle{{R}e{F}in{ED}: An Efficient Zero-shot-capable
  Approach to End-to-End Entity Linking}. In
  \bibinfo{booktitle}{\emph{Proceedings of the 2022 Conference of the North
  American Chapter of the Association for Computational Linguistics: Human
  Language Technologies: Industry Track}}. \bibinfo{publisher}{Association for
  Computational Linguistics}, \bibinfo{address}{Hybrid: Seattle, Washington +
  Online}, \bibinfo{pages}{209--220}.
\newblock
\urldef\tempurl%
\url{https://doi.org/10.18653/v1/2022.naacl-industry.24}
\showDOI{\tempurl}


\bibitem[Basaldella et~al\mbox{.}(2020)]%
        {basaldella2020cometa}
\bibfield{author}{\bibinfo{person}{Marco Basaldella}, \bibinfo{person}{Fangyu
  Liu}, \bibinfo{person}{Ehsan Shareghi}, {and} \bibinfo{person}{Nigel
  Collier}.} \bibinfo{year}{2020}\natexlab{}.
\newblock \showarticletitle{{COMETA}: A Corpus for Medical Entity Linking in
  the Social Media}. In \bibinfo{booktitle}{\emph{Proceedings of the 2020
  Conference on Empirical Methods in Natural Language Processing (EMNLP)}}.
  \bibinfo{publisher}{Association for Computational Linguistics},
  \bibinfo{address}{Online}, \bibinfo{pages}{3122--3137}.
\newblock
\urldef\tempurl%
\url{https://doi.org/10.18653/v1/2020.emnlp-main.253}
\showDOI{\tempurl}


\bibitem[Bhowmik et~al\mbox{.}(2021)]%
        {bhowmik2021}
\bibfield{author}{\bibinfo{person}{Rajarshi Bhowmik}, \bibinfo{person}{Karl
  Stratos}, {and} \bibinfo{person}{Gerard de Melo}.}
  \bibinfo{year}{2021}\natexlab{}.
\newblock \showarticletitle{Fast and Effective Biomedical Entity Linking Using
  a Dual Encoder}. In \bibinfo{booktitle}{\emph{Proceedings of the 12th
  International Workshop on Health Text Mining and Information Analysis}}.
  \bibinfo{publisher}{Association for Computational Linguistics},
  \bibinfo{address}{online}, \bibinfo{pages}{28--37}.
\newblock
\urldef\tempurl%
\url{https://aclanthology.org/2021.louhi-1.4}
\showURL{%
\tempurl}


\bibitem[Bodenreider(2004)]%
        {Bodenreider2004umls}
\bibfield{author}{\bibinfo{person}{Olivier Bodenreider}.}
  \bibinfo{year}{2004}\natexlab{}.
\newblock \showarticletitle{{The Unified Medical Language System (UMLS):
  integrating biomedical terminology}}.
\newblock \bibinfo{journal}{\emph{Nucleic Acids Research}}
  \bibinfo{volume}{32}, \bibinfo{number}{suppl\_1} (\bibinfo{date}{01}
  \bibinfo{year}{2004}), \bibinfo{pages}{D267--D270}.
\newblock
\showISSN{0305-1048}
\urldef\tempurl%
\url{https://doi.org/10.1093/nar/gkh061}
\showDOI{\tempurl}


\bibitem[Bunescu and Pa{\c{s}}ca(2006)]%
        {bunescu-pasca-2006}
\bibfield{author}{\bibinfo{person}{Razvan Bunescu} {and}
  \bibinfo{person}{Marius Pa{\c{s}}ca}.} \bibinfo{year}{2006}\natexlab{}.
\newblock \showarticletitle{Using Encyclopedic Knowledge for Named entity
  Disambiguation}. In \bibinfo{booktitle}{\emph{11th Conference of the
  {E}uropean Chapter of the Association for Computational Linguistics}}.
  \bibinfo{publisher}{Association for Computational Linguistics},
  \bibinfo{address}{Trento, Italy}, \bibinfo{pages}{9--16}.
\newblock
\urldef\tempurl%
\url{https://aclanthology.org/E06-1002}
\showURL{%
\tempurl}


\bibitem[Chen et~al\mbox{.}(2021)]%
        {chen2021lightweight}
\bibfield{author}{\bibinfo{person}{Lihu Chen}, \bibinfo{person}{Gaël
  Varoquaux}, {and} \bibinfo{person}{Fabian~M. Suchanek}.}
  \bibinfo{year}{2021}\natexlab{}.
\newblock \showarticletitle{A Lightweight Neural Model for Biomedical Entity
  Linking}.
\newblock \bibinfo{journal}{\emph{Proceedings of the AAAI Conference on
  Artificial Intelligence}} \bibinfo{volume}{35}, \bibinfo{number}{14}
  (\bibinfo{date}{May} \bibinfo{year}{2021}), \bibinfo{pages}{12657--12665}.
\newblock
\urldef\tempurl%
\url{https://doi.org/10.1609/aaai.v35i14.17499}
\showDOI{\tempurl}


\bibitem[Devlin et~al\mbox{.}(2019)]%
        {devlin-etal-2019-bert}
\bibfield{author}{\bibinfo{person}{Jacob Devlin}, \bibinfo{person}{Ming-Wei
  Chang}, \bibinfo{person}{Kenton Lee}, {and} \bibinfo{person}{Kristina
  Toutanova}.} \bibinfo{year}{2019}\natexlab{}.
\newblock \showarticletitle{{BERT}: Pre-training of Deep Bidirectional
  Transformers for Language Understanding}. In
  \bibinfo{booktitle}{\emph{Proceedings of the 2019 Conference of the North
  {A}merican Chapter of the Association for Computational Linguistics: Human
  Language Technologies, Volume 1 (Long and Short Papers)}}.
  \bibinfo{publisher}{Association for Computational Linguistics},
  \bibinfo{address}{Minneapolis, Minnesota}, \bibinfo{pages}{4171--4186}.
\newblock
\urldef\tempurl%
\url{https://doi.org/10.18653/v1/N19-1423}
\showDOI{\tempurl}


\bibitem[Donnelly et~al\mbox{.}(2006)]%
        {donnelly2006snomed}
\bibfield{author}{\bibinfo{person}{Kevin Donnelly} {et~al\mbox{.}}}
  \bibinfo{year}{2006}\natexlab{}.
\newblock \showarticletitle{{SNOMED-CT}: The advanced terminology and coding
  system for eHealth}.
\newblock In \bibinfo{booktitle}{\emph{Medical and Care Compunetics 3}}.
  \bibinfo{series}{Studies in health technology and informatics},
  Vol.~\bibinfo{volume}{121}. \bibinfo{publisher}{IOS Press},
  \bibinfo{address}{Amsterdam, Netherlands}, \bibinfo{pages}{279--290}.
\newblock


\bibitem[Dredze et~al\mbox{.}(2010)]%
        {dredze2010tac}
\bibfield{author}{\bibinfo{person}{Mark Dredze}, \bibinfo{person}{Paul
  McNamee}, \bibinfo{person}{Delip Rao}, \bibinfo{person}{Adam Gerber}, {and}
  \bibinfo{person}{Tim Finin}.} \bibinfo{year}{2010}\natexlab{}.
\newblock \showarticletitle{Entity Disambiguation for Knowledge Base
  Population}. In \bibinfo{booktitle}{\emph{Proceedings of the 23rd
  International Conference on Computational Linguistics (Coling 2010)}}.
  \bibinfo{publisher}{Coling 2010 Organizing Committee},
  \bibinfo{address}{Beijing, China}, \bibinfo{pages}{277--285}.
\newblock
\urldef\tempurl%
\url{https://aclanthology.org/C10-1032}
\showURL{%
\tempurl}


\bibitem[D{'}Souza and Ng(2015)]%
        {dsouza-ng-2015-sieve-based}
\bibfield{author}{\bibinfo{person}{Jennifer D{'}Souza} {and}
  \bibinfo{person}{Vincent Ng}.} \bibinfo{year}{2015}\natexlab{}.
\newblock \showarticletitle{Sieve-Based Entity Linking for the Biomedical
  Domain}. In \bibinfo{booktitle}{\emph{Proceedings of the 53rd Annual Meeting
  of the Association for Computational Linguistics and the 7th International
  Joint Conference on Natural Language Processing (Volume 2: Short Papers)}}.
  \bibinfo{publisher}{Association for Computational Linguistics},
  \bibinfo{address}{Beijing, China}, \bibinfo{pages}{297--302}.
\newblock
\urldef\tempurl%
\url{https://doi.org/10.3115/v1/P15-2049}
\showDOI{\tempurl}


\bibitem[Ehrmann et~al\mbox{.}(2020)]%
        {ehrmann2020extended}
\bibfield{author}{\bibinfo{person}{Maud Ehrmann}, \bibinfo{person}{Matteo
  Romanello}, \bibinfo{person}{Alex Fl{\"u}ckiger}, {and}
  \bibinfo{person}{Simon Clematide}.} \bibinfo{year}{2020}\natexlab{}.
\newblock \showarticletitle{Extended overview of CLEF HIPE 2020: named entity
  processing on historical newspapers}. In \bibinfo{booktitle}{\emph{Working
  Notes of CLEF 2020 - Conference and Labs of the Evaluation Forum}}.
  \bibinfo{publisher}{CEUR Workshop Proceedings (CEUR-WS.org)},
  \bibinfo{address}{Thessaloniki, Greece}, \bibinfo{pages}{1--38}.
\newblock


\bibitem[Elhadad et~al\mbox{.}(2013)]%
        {shareclef2013guideline}
\bibfield{author}{\bibinfo{person}{Noémie Elhadad}, \bibinfo{person}{Wendy
  Chapman}, {and} \bibinfo{person}{Guergana Savova}.}
  \bibinfo{year}{2013}\natexlab{}.
\newblock \bibinfo{title}{ShAReCLEF eHealth 2013: Natural Language Processing
  and Information Retrieval for Clinical Care 1.0}.
\newblock
  \bibinfo{howpublished}{\url{https://physionet.org/content/shareclefehealth2013/1.0/Task1ShAReGuidelines2013.pdf}}.
\newblock


\bibitem[Gao et~al\mbox{.}(2021)]%
        {gao-etal-2021-simcse}
\bibfield{author}{\bibinfo{person}{Tianyu Gao}, \bibinfo{person}{Xingcheng
  Yao}, {and} \bibinfo{person}{Danqi Chen}.} \bibinfo{year}{2021}\natexlab{}.
\newblock \showarticletitle{{S}im{CSE}: Simple Contrastive Learning of Sentence
  Embeddings}. In \bibinfo{booktitle}{\emph{Proceedings of the 2021 Conference
  on Empirical Methods in Natural Language Processing}}.
  \bibinfo{publisher}{Association for Computational Linguistics},
  \bibinfo{address}{Online and Punta Cana, Dominican Republic},
  \bibinfo{pages}{6894--6910}.
\newblock
\urldef\tempurl%
\url{https://doi.org/10.18653/v1/2021.emnlp-main.552}
\showDOI{\tempurl}


\bibitem[Gruber(1995)]%
        {GRUBER1995907}
\bibfield{author}{\bibinfo{person}{Thomas~R. Gruber}.}
  \bibinfo{year}{1995}\natexlab{}.
\newblock \showarticletitle{Toward principles for the design of ontologies used
  for knowledge sharing?}
\newblock \bibinfo{journal}{\emph{International Journal of Human-Computer
  Studies}} \bibinfo{volume}{43}, \bibinfo{number}{5} (\bibinfo{year}{1995}),
  \bibinfo{pages}{907--928}.
\newblock
\showISSN{1071-5819}
\urldef\tempurl%
\url{https://doi.org/10.1006/ijhc.1995.1081}
\showDOI{\tempurl}


\bibitem[Gu et~al\mbox{.}(2021a)]%
        {gu2021read}
\bibfield{author}{\bibinfo{person}{Yingjie Gu}, \bibinfo{person}{Xiaoye Qu},
  \bibinfo{person}{Zhefeng Wang}, \bibinfo{person}{Baoxing Huai},
  \bibinfo{person}{Nicholas~Jing Yuan}, {and} \bibinfo{person}{Xiaolin Gui}.}
  \bibinfo{year}{2021}\natexlab{a}.
\newblock \showarticletitle{Read, Retrospect, Select: An MRC Framework to Short
  Text Entity Linking}.
\newblock \bibinfo{journal}{\emph{Proceedings of the AAAI Conference on
  Artificial Intelligence}} \bibinfo{volume}{35}, \bibinfo{number}{14}
  (\bibinfo{date}{May} \bibinfo{year}{2021}), \bibinfo{pages}{12920--12928}.
\newblock
\urldef\tempurl%
\url{https://doi.org/10.1609/aaai.v35i14.17528}
\showDOI{\tempurl}


\bibitem[Gu et~al\mbox{.}(2021b)]%
        {gu2022pmbert}
\bibfield{author}{\bibinfo{person}{Yu Gu}, \bibinfo{person}{Robert Tinn},
  \bibinfo{person}{Hao Cheng}, \bibinfo{person}{Michael Lucas},
  \bibinfo{person}{Naoto Usuyama}, \bibinfo{person}{Xiaodong Liu},
  \bibinfo{person}{Tristan Naumann}, \bibinfo{person}{Jianfeng Gao}, {and}
  \bibinfo{person}{Hoifung Poon}.} \bibinfo{year}{2021}\natexlab{b}.
\newblock \showarticletitle{Domain-Specific Language Model Pretraining for
  Biomedical Natural Language Processing}.
\newblock \bibinfo{journal}{\emph{ACM Trans. Comput. Healthcare}}
  \bibinfo{volume}{3}, \bibinfo{number}{1}, Article \bibinfo{articleno}{2}
  (\bibinfo{date}{oct} \bibinfo{year}{2021}), \bibinfo{numpages}{23}~pages.
\newblock
\showISSN{2691-1957}
\urldef\tempurl%
\url{https://doi.org/10.1145/3458754}
\showDOI{\tempurl}


\bibitem[He et~al\mbox{.}(2023)]%
        {he2023deeponto}
\bibfield{author}{\bibinfo{person}{Yuan He}, \bibinfo{person}{Jiaoyan Chen},
  \bibinfo{person}{Hang Dong}, \bibinfo{person}{Ian Horrocks},
  \bibinfo{person}{Carlo Allocca}, \bibinfo{person}{Taehun Kim}, {and}
  \bibinfo{person}{Brahmananda Sapkota}.} \bibinfo{year}{2023}\natexlab{}.
\newblock \showarticletitle{DeepOnto: A Python Package for Ontology Engineering
  with Deep Learning}.
\newblock \bibinfo{journal}{\emph{arXiv preprint arXiv:2307.03067}}
  (\bibinfo{year}{2023}).
\newblock


\bibitem[He et~al\mbox{.}(2022)]%
        {yuan2022resource}
\bibfield{author}{\bibinfo{person}{Yuan He}, \bibinfo{person}{Jiaoyan Chen},
  \bibinfo{person}{Hang Dong}, \bibinfo{person}{Ernesto Jim{\'e}nez-Ruiz},
  \bibinfo{person}{Ali Hadian}, {and} \bibinfo{person}{Ian Horrocks}.}
  \bibinfo{year}{2022}\natexlab{}.
\newblock \showarticletitle{Machine learning-friendly biomedical datasets for
  equivalence and subsumption ontology matching}. In
  \bibinfo{booktitle}{\emph{International Semantic Web Conference}}.
  \bibinfo{publisher}{Springer}, \bibinfo{address}{Cham, Switzerland},
  \bibinfo{pages}{575--591}.
\newblock


\bibitem[Heist and Paulheim(2023)]%
        {heist2023nastylinker}
\bibfield{author}{\bibinfo{person}{Nicolas Heist} {and} \bibinfo{person}{Heiko
  Paulheim}.} \bibinfo{year}{2023}\natexlab{}.
\newblock \showarticletitle{NASTyLinker: NIL-Aware Scalable Transformer-Based
  Entity Linker}. In \bibinfo{booktitle}{\emph{The Semantic Web - 20th
  International Conference, {ESWC} 2023, Hersonissos, Crete, Greece, May 28 -
  June 1, 2023, Proceedings}} \emph{(\bibinfo{series}{Lecture Notes in Computer
  Science}, Vol.~\bibinfo{volume}{13870})},
  \bibfield{editor}{\bibinfo{person}{Catia Pesquita}, \bibinfo{person}{Ernesto
  Jim{\'{e}}nez{-}Ruiz}, \bibinfo{person}{Jamie~P. McCusker},
  \bibinfo{person}{Daniel Faria}, \bibinfo{person}{Mauro Dragoni},
  \bibinfo{person}{Anastasia Dimou}, \bibinfo{person}{Rapha{\"{e}}l Troncy},
  {and} \bibinfo{person}{Sven Hertling}} (Eds.). \bibinfo{publisher}{Springer},
  \bibinfo{address}{Cham}, \bibinfo{pages}{174--191}.
\newblock


\bibitem[Hoffart et~al\mbox{.}(2014)]%
        {Hoffart2014}
\bibfield{author}{\bibinfo{person}{Johannes Hoffart}, \bibinfo{person}{Yasemin
  Altun}, {and} \bibinfo{person}{Gerhard Weikum}.}
  \bibinfo{year}{2014}\natexlab{}.
\newblock \showarticletitle{Discovering Emerging Entities with Ambiguous
  Names}. In \bibinfo{booktitle}{\emph{Proceedings of the 23rd International
  Conference on World Wide Web}} (Seoul, Korea) \emph{(\bibinfo{series}{WWW
  '14})}. \bibinfo{publisher}{Association for Computing Machinery},
  \bibinfo{address}{New York, NY, USA}, \bibinfo{pages}{385–396}.
\newblock
\showISBNx{9781450327442}
\urldef\tempurl%
\url{https://doi.org/10.1145/2566486.2568003}
\showDOI{\tempurl}


\bibitem[Hoffart et~al\mbox{.}(2016)]%
        {Hoffart2016}
\bibfield{author}{\bibinfo{person}{Johannes Hoffart}, \bibinfo{person}{Dragan
  Milchevski}, \bibinfo{person}{Gerhard Weikum}, \bibinfo{person}{Avishek
  Anand}, {and} \bibinfo{person}{Jaspreet Singh}.}
  \bibinfo{year}{2016}\natexlab{}.
\newblock \showarticletitle{The Knowledge Awakens: Keeping Knowledge Bases
  Fresh with Emerging Entities}. In \bibinfo{booktitle}{\emph{Proceedings of
  the 25th International Conference Companion on World Wide Web}}
  (Montr\'{e}al, Qu\'{e}bec, Canada) \emph{(\bibinfo{series}{WWW '16
  Companion})}. \bibinfo{publisher}{International World Wide Web Conferences
  Steering Committee}, \bibinfo{address}{Republic and Canton of Geneva, CHE},
  \bibinfo{pages}{203–206}.
\newblock
\showISBNx{9781450341448}
\urldef\tempurl%
\url{https://doi.org/10.1145/2872518.2890537}
\showDOI{\tempurl}


\bibitem[Iurshina et~al\mbox{.}(2022)]%
        {Iurshina2022nilk}
\bibfield{author}{\bibinfo{person}{Anastasiia Iurshina},
  \bibinfo{person}{Jiaxin Pan}, \bibinfo{person}{Rafika Boutalbi}, {and}
  \bibinfo{person}{Steffen Staab}.} \bibinfo{year}{2022}\natexlab{}.
\newblock \showarticletitle{NILK: Entity Linking Dataset Targeting NIL-Linking
  Cases}. In \bibinfo{booktitle}{\emph{Proceedings of the 31st ACM
  International Conference on Information \& Knowledge Management}} (Atlanta,
  GA, USA) \emph{(\bibinfo{series}{CIKM '22})}. \bibinfo{publisher}{Association
  for Computing Machinery}, \bibinfo{address}{New York, NY, USA},
  \bibinfo{pages}{4069–4073}.
\newblock
\showISBNx{9781450392365}
\urldef\tempurl%
\url{https://doi.org/10.1145/3511808.3557659}
\showDOI{\tempurl}


\bibitem[Ji et~al\mbox{.}(2011)]%
        {ji2011TAC}
\bibfield{author}{\bibinfo{person}{Heng Ji}, \bibinfo{person}{Ralph Grishman},
  \bibinfo{person}{Hoa~Trang Dang}, \bibinfo{person}{Kira Griffitt}, {and}
  \bibinfo{person}{Joe Ellis}.} \bibinfo{year}{2011}\natexlab{}.
\newblock \showarticletitle{Overview of the TAC 2011 knowledge base population
  track}. In \bibinfo{booktitle}{\emph{Third text analysis conference (TAC
  2011)}}. \bibinfo{publisher}{National Institute of Standards and Technology},
  \bibinfo{address}{Gaithersburg, Maryland, USA}, \bibinfo{pages}{1--33}.
\newblock


\bibitem[Ji et~al\mbox{.}(2020)]%
        {ji2020bert}
\bibfield{author}{\bibinfo{person}{Zongcheng Ji}, \bibinfo{person}{Qiang Wei},
  {and} \bibinfo{person}{Hua Xu}.} \bibinfo{year}{2020}\natexlab{}.
\newblock \showarticletitle{{BERT}-based ranking for biomedical entity
  normalization}.
\newblock \bibinfo{journal}{\emph{AMIA Summits on Translational Science
  Proceedings}}  \bibinfo{volume}{2020} (\bibinfo{year}{2020}),
  \bibinfo{pages}{269}.
\newblock


\bibitem[Kassner et~al\mbox{.}(2022)]%
        {kassner2022edin}
\bibfield{author}{\bibinfo{person}{Nora Kassner}, \bibinfo{person}{Fabio
  Petroni}, \bibinfo{person}{Mikhail Plekhanov}, \bibinfo{person}{Sebastian
  Riedel}, {and} \bibinfo{person}{Nicola Cancedda}.}
  \bibinfo{year}{2022}\natexlab{}.
\newblock \showarticletitle{{EDIN}: An End-to-end Benchmark and Pipeline for
  Unknown Entity Discovery and Indexing}. In
  \bibinfo{booktitle}{\emph{Proceedings of the 2022 Conference on Empirical
  Methods in Natural Language Processing}}. \bibinfo{publisher}{Association for
  Computational Linguistics}, \bibinfo{address}{Abu Dhabi, United Arab
  Emirates}, \bibinfo{pages}{8659--8673}.
\newblock


\bibitem[Liu et~al\mbox{.}(2021)]%
        {liu2021sap}
\bibfield{author}{\bibinfo{person}{Fangyu Liu}, \bibinfo{person}{Ehsan
  Shareghi}, \bibinfo{person}{Zaiqiao Meng}, \bibinfo{person}{Marco
  Basaldella}, {and} \bibinfo{person}{Nigel Collier}.}
  \bibinfo{year}{2021}\natexlab{}.
\newblock \showarticletitle{Self-Alignment Pretraining for Biomedical Entity
  Representations}. In \bibinfo{booktitle}{\emph{Proceedings of the 2021
  Conference of the North American Chapter of the Association for Computational
  Linguistics: Human Language Technologies}}. \bibinfo{publisher}{Association
  for Computational Linguistics}, \bibinfo{address}{Online},
  \bibinfo{pages}{4228--4238}.
\newblock
\urldef\tempurl%
\url{https://doi.org/10.18653/v1/2021.naacl-main.334}
\showDOI{\tempurl}


\bibitem[Loshchilov and Hutter(2019)]%
        {loshchilov2018decoupled}
\bibfield{author}{\bibinfo{person}{Ilya Loshchilov} {and}
  \bibinfo{person}{Frank Hutter}.} \bibinfo{year}{2019}\natexlab{}.
\newblock \showarticletitle{Decoupled Weight Decay Regularization}. In
  \bibinfo{booktitle}{\emph{7th International Conference on Learning
  Representations, {ICLR} 2019}}. \bibinfo{publisher}{OpenReview.net},
  \bibinfo{address}{New Orleans, LA, USA}, \bibinfo{pages}{1--10}.
\newblock


\bibitem[Luo et~al\mbox{.}(2020)]%
        {Luo2020}
\bibfield{author}{\bibinfo{person}{Yen-Fu Luo}, \bibinfo{person}{Sam Henry},
  \bibinfo{person}{Yanshan Wang}, \bibinfo{person}{Feichen Shen},
  \bibinfo{person}{Ozlem Uzuner}, {and} \bibinfo{person}{Anna Rumshisky}.}
  \bibinfo{year}{2020}\natexlab{}.
\newblock \showarticletitle{{The 2019 n2c2/UMass Lowell shared task on clinical
  concept normalization}}.
\newblock \bibinfo{journal}{\emph{Journal of the American Medical Informatics
  Association}} \bibinfo{volume}{27}, \bibinfo{number}{10} (\bibinfo{date}{09}
  \bibinfo{year}{2020}), \bibinfo{pages}{1529--e1}.
\newblock
\showISSN{1527-974X}
\urldef\tempurl%
\url{https://doi.org/10.1093/jamia/ocaa106}
\showDOI{\tempurl}


\bibitem[Martins et~al\mbox{.}(2019)]%
        {martins-etal-2019-joint}
\bibfield{author}{\bibinfo{person}{Pedro~Henrique Martins},
  \bibinfo{person}{Zita Marinho}, {and} \bibinfo{person}{Andr{\'e} F.~T.
  Martins}.} \bibinfo{year}{2019}\natexlab{}.
\newblock \showarticletitle{Joint Learning of Named Entity Recognition and
  Entity Linking}. In \bibinfo{booktitle}{\emph{Proceedings of the 57th Annual
  Meeting of the Association for Computational Linguistics: Student Research
  Workshop}}. \bibinfo{publisher}{Association for Computational Linguistics},
  \bibinfo{address}{Florence, Italy}, \bibinfo{pages}{190--196}.
\newblock
\urldef\tempurl%
\url{https://doi.org/10.18653/v1/P19-2026}
\showDOI{\tempurl}


\bibitem[McNamee et~al\mbox{.}(2009)]%
        {mcnamee2009hltcoe}
\bibfield{author}{\bibinfo{person}{Paul McNamee}, \bibinfo{person}{Mark
  Dredze}, \bibinfo{person}{Adam Gerber}, \bibinfo{person}{Nikesh Garera},
  \bibinfo{person}{Tim Finin}, \bibinfo{person}{James Mayfield},
  \bibinfo{person}{Christine Piatko}, \bibinfo{person}{Delip Rao},
  \bibinfo{person}{David Yarowsky}, \bibinfo{person}{Markus Dreyer},
  {et~al\mbox{.}}} \bibinfo{year}{2009}\natexlab{}.
\newblock \showarticletitle{HLTCOE Approaches to Knowledge Base Population at
  TAC 2009}. In \bibinfo{booktitle}{\emph{Proceedings of the 2009 Text Analysis
  Conference}}. \bibinfo{publisher}{National Institute of Standards and
  Technology}, \bibinfo{address}{Gaithersburg, Maryland, USA},
  \bibinfo{pages}{1--10}.
\newblock


\bibitem[Mohan and Li(2019)]%
        {mohan2019medmentions}
\bibfield{author}{\bibinfo{person}{Sunil Mohan} {and} \bibinfo{person}{Donghui
  Li}.} \bibinfo{year}{2019}\natexlab{}.
\newblock \showarticletitle{MedMentions: A Large Biomedical Corpus Annotated
  with {\{}UMLS{\}} Concepts}. In \bibinfo{booktitle}{\emph{Automated Knowledge
  Base Construction (AKBC)}}. \bibinfo{publisher}{openreview.net},
  \bibinfo{address}{Amherst, MA, USA}, \bibinfo{pages}{1--13}.
\newblock
\urldef\tempurl%
\url{https://doi.org/10.24432/C5G59C}
\showDOI{\tempurl}


\bibitem[M{\"{o}}ller(2022)]%
        {moller2022knowledge}
\bibfield{author}{\bibinfo{person}{Cedric M{\"{o}}ller}.}
  \bibinfo{year}{2022}\natexlab{}.
\newblock \showarticletitle{Knowledge Graph Population with Out-of-KG
  Entities}. In \bibinfo{booktitle}{\emph{The Semantic Web: {ESWC} 2022
  Satellite Events - Hersonissos}} \emph{(\bibinfo{series}{Lecture Notes in
  Computer Science}, Vol.~\bibinfo{volume}{13384})},
  \bibfield{editor}{\bibinfo{person}{Paul Groth}, \bibinfo{person}{Anisa Rula},
  \bibinfo{person}{Jodi Schneider}, \bibinfo{person}{Ilaria Tiddi},
  \bibinfo{person}{Elena Simperl}, \bibinfo{person}{Panos Alexopoulos},
  \bibinfo{person}{Rinke Hoekstra}, \bibinfo{person}{Mehwish Alam},
  \bibinfo{person}{Anastasia Dimou}, {and} \bibinfo{person}{Minna Tamper}}
  (Eds.). \bibinfo{publisher}{Springer}, \bibinfo{address}{Hersonissos, Crete,
  Greece}, \bibinfo{pages}{199--214}.
\newblock
\urldef\tempurl%
\url{https://doi.org/10.1007/978-3-031-11609-4\_35}
\showDOI{\tempurl}


\bibitem[Rao et~al\mbox{.}(2013)]%
        {poibeau_entity_2013}
\bibfield{author}{\bibinfo{person}{Delip Rao}, \bibinfo{person}{Paul McNamee},
  {and} \bibinfo{person}{Mark Dredze}.} \bibinfo{year}{2013}\natexlab{}.
\newblock \showarticletitle{Entity {Linking}: {Finding} {Extracted} {Entities}
  in a {Knowledge} {Base}}.
\newblock In \bibinfo{booktitle}{\emph{Multi-source, {Multilingual}
  {Information} {Extraction} and {Summarization}}},
  \bibfield{editor}{\bibinfo{person}{Thierry Poibeau}, \bibinfo{person}{Horacio
  Saggion}, \bibinfo{person}{Jakub Piskorski}, {and} \bibinfo{person}{Roman
  Yangarber}} (Eds.). \bibinfo{publisher}{Springer Berlin Heidelberg},
  \bibinfo{address}{Berlin, Heidelberg}, \bibinfo{pages}{93--115}.
\newblock
\showISBNx{978-3-642-28568-4 978-3-642-28569-1}
\urldef\tempurl%
\url{https://doi.org/10.1007/978-3-642-28569-1_5}
\showDOI{\tempurl}
\newblock
\shownote{Series Title: Theory and Applications of Natural Language
  Processing}.


\bibitem[Reimers and Gurevych(2019)]%
        {reimers-gurevych-2019-sentence}
\bibfield{author}{\bibinfo{person}{Nils Reimers} {and} \bibinfo{person}{Iryna
  Gurevych}.} \bibinfo{year}{2019}\natexlab{}.
\newblock \showarticletitle{Sentence-{BERT}: Sentence Embeddings using
  {S}iamese {BERT}-Networks}. In \bibinfo{booktitle}{\emph{Proceedings of the
  2019 Conference on Empirical Methods in Natural Language Processing and the
  9th International Joint Conference on Natural Language Processing
  (EMNLP-IJCNLP)}}. \bibinfo{publisher}{Association for Computational
  Linguistics}, \bibinfo{address}{Hong Kong, China},
  \bibinfo{pages}{3982--3992}.
\newblock
\urldef\tempurl%
\url{https://doi.org/10.18653/v1/D19-1410}
\showDOI{\tempurl}


\bibitem[Ristoski et~al\mbox{.}(2021)]%
        {Ristoski2021kgzselEbay}
\bibfield{author}{\bibinfo{person}{Petar Ristoski}, \bibinfo{person}{Zhizhong
  Lin}, {and} \bibinfo{person}{Qunzhi Zhou}.} \bibinfo{year}{2021}\natexlab{}.
\newblock \showarticletitle{KG-ZESHEL: Knowledge Graph-Enhanced Zero-Shot
  Entity Linking}. In \bibinfo{booktitle}{\emph{Proceedings of the 11th on
  Knowledge Capture Conference}} (Virtual Event, USA)
  \emph{(\bibinfo{series}{K-CAP '21})}. \bibinfo{publisher}{Association for
  Computing Machinery}, \bibinfo{address}{New York, NY, USA},
  \bibinfo{pages}{49–56}.
\newblock
\showISBNx{9781450384575}
\urldef\tempurl%
\url{https://doi.org/10.1145/3460210.3493549}
\showDOI{\tempurl}


\bibitem[Rizzo et~al\mbox{.}(2017)]%
        {rizzo2017NEEL}
\bibfield{author}{\bibinfo{person}{Giuseppe Rizzo}, \bibinfo{person}{Bianca
  Pereira}, \bibinfo{person}{Andrea Varga}, \bibinfo{person}{Marieke Van~Erp},
  {and} \bibinfo{person}{Amparo~Elizabeth Cano~Basave}.}
  \bibinfo{year}{2017}\natexlab{}.
\newblock \showarticletitle{Lessons learnt from the Named Entity rEcognition
  and Linking (NEEL) challenge series}.
\newblock \bibinfo{journal}{\emph{Semantic Web}} \bibinfo{volume}{8},
  \bibinfo{number}{5} (\bibinfo{year}{2017}), \bibinfo{pages}{667--700}.
\newblock


\bibitem[Sevgili et~al\mbox{.}(2022)]%
        {sevgili_neural_2022}
\bibfield{author}{\bibinfo{person}{\"{O}zge Sevgili}, \bibinfo{person}{Artem
  Shelmanov}, \bibinfo{person}{Mikhail Arkhipov}, \bibinfo{person}{Alexander
  Panchenko}, {and} \bibinfo{person}{Chris Biemann}.}
  \bibinfo{year}{2022}\natexlab{}.
\newblock \showarticletitle{Neural entity linking: {A} survey of models based
  on deep learning}.
\newblock \bibinfo{journal}{\emph{Semantic Web}} \bibinfo{volume}{13},
  \bibinfo{number}{3} (\bibinfo{date}{Jan.} \bibinfo{year}{2022}),
  \bibinfo{pages}{527--570}.
\newblock
\showISSN{1570-0844}
\urldef\tempurl%
\url{https://doi.org/10.3233/SW-222986}
\showDOI{\tempurl}
\newblock
\shownote{Publisher: IOS Press}.


\bibitem[Shen et~al\mbox{.}(2023)]%
        {shen2021eldeep}
\bibfield{author}{\bibinfo{person}{Wei Shen}, \bibinfo{person}{Yuhan Li},
  \bibinfo{person}{Yinan Liu}, \bibinfo{person}{Jiawei Han},
  \bibinfo{person}{Jianyong Wang}, {and} \bibinfo{person}{Xiaojie Yuan}.}
  \bibinfo{year}{2023}\natexlab{}.
\newblock \showarticletitle{Entity Linking Meets Deep Learning: Techniques and
  Solutions}.
\newblock \bibinfo{journal}{\emph{IEEE Transactions on Knowledge and Data
  Engineering}} \bibinfo{volume}{35}, \bibinfo{number}{3}
  (\bibinfo{year}{2023}), \bibinfo{pages}{2556--2578}.
\newblock
\urldef\tempurl%
\url{https://doi.org/10.1109/TKDE.2021.3117715}
\showDOI{\tempurl}


\bibitem[Shen et~al\mbox{.}(2014)]%
        {shen2014entity}
\bibfield{author}{\bibinfo{person}{Wei Shen}, \bibinfo{person}{Jianyong Wang},
  {and} \bibinfo{person}{Jiawei Han}.} \bibinfo{year}{2014}\natexlab{}.
\newblock \showarticletitle{Entity linking with a knowledge base: Issues,
  techniques, and solutions}.
\newblock \bibinfo{journal}{\emph{IEEE Transactions on Knowledge and Data
  Engineering}} \bibinfo{volume}{27}, \bibinfo{number}{2}
  (\bibinfo{year}{2014}), \bibinfo{pages}{443--460}.
\newblock


\bibitem[Suominen et~al\mbox{.}(2013)]%
        {suominen2013shareclef}
\bibfield{author}{\bibinfo{person}{Hanna Suominen}, \bibinfo{person}{Sanna
  Salanter{\"{a}}}, \bibinfo{person}{Sumithra Velupillai},
  \bibinfo{person}{Wendy~Webber Chapman}, \bibinfo{person}{Guergana~K. Savova},
  \bibinfo{person}{Noemie Elhadad}, \bibinfo{person}{Sameer Pradhan},
  \bibinfo{person}{Brett~R. South}, \bibinfo{person}{Danielle~L. Mowery},
  \bibinfo{person}{Gareth J.~F. Jones}, \bibinfo{person}{Johannes Leveling},
  \bibinfo{person}{Liadh Kelly}, \bibinfo{person}{Lorraine Goeuriot},
  \bibinfo{person}{David Mart{\'{\i}}nez}, {and} \bibinfo{person}{Guido
  Zuccon}.} \bibinfo{year}{2013}\natexlab{}.
\newblock \showarticletitle{Overview of the ShARe/CLEF eHealth Evaluation Lab
  2013}. In \bibinfo{booktitle}{\emph{Information Access Evaluation.
  Multilinguality, Multimodality, and Visualization - 4th International
  Conference of the {CLEF} Initiative, {CLEF} 2013}}
  \emph{(\bibinfo{series}{Lecture Notes in Computer Science},
  Vol.~\bibinfo{volume}{8138})}, \bibfield{editor}{\bibinfo{person}{Pamela
  Forner}, \bibinfo{person}{Henning M{\"{u}}ller}, \bibinfo{person}{Roberto
  Paredes}, \bibinfo{person}{Paolo Rosso}, {and} \bibinfo{person}{Benno Stein}}
  (Eds.). \bibinfo{publisher}{Springer}, \bibinfo{address}{Valencia, Spain},
  \bibinfo{pages}{212--231}.
\newblock
\urldef\tempurl%
\url{https://doi.org/10.1007/978-3-642-40802-1\_24}
\showDOI{\tempurl}


\bibitem[Twigg et~al\mbox{.}(2016)]%
        {twigg2016recurrent}
\bibfield{author}{\bibinfo{person}{Stephen~RF Twigg}, \bibinfo{person}{Robert~B
  Hufnagel}, \bibinfo{person}{Kerry~A Miller}, \bibinfo{person}{Yan Zhou},
  \bibinfo{person}{Simon~J McGowan}, \bibinfo{person}{John Taylor},
  \bibinfo{person}{Jude Craft}, \bibinfo{person}{Jenny~C Taylor},
  \bibinfo{person}{Stephanie~L Santoro}, \bibinfo{person}{Taosheng Huang},
  {et~al\mbox{.}}} \bibinfo{year}{2016}\natexlab{}.
\newblock \showarticletitle{A recurrent mosaic mutation in SMO, encoding the
  hedgehog signal transducer smoothened, is the major cause of Curry-Jones
  syndrome}.
\newblock \bibinfo{journal}{\emph{The American Journal of Human Genetics}}
  \bibinfo{volume}{98}, \bibinfo{number}{6} (\bibinfo{year}{2016}),
  \bibinfo{pages}{1256--1265}.
\newblock


\bibitem[Vrande\v{c}i\'{c} and Kr\"{o}tzsch(2014)]%
        {vrandecic2014wikidata}
\bibfield{author}{\bibinfo{person}{Denny Vrande\v{c}i\'{c}} {and}
  \bibinfo{person}{Markus Kr\"{o}tzsch}.} \bibinfo{year}{2014}\natexlab{}.
\newblock \showarticletitle{Wikidata: A Free Collaborative Knowledgebase}.
\newblock \bibinfo{journal}{\emph{Commun. ACM}} \bibinfo{volume}{57},
  \bibinfo{number}{10} (\bibinfo{date}{sep} \bibinfo{year}{2014}),
  \bibinfo{pages}{78–85}.
\newblock
\showISSN{0001-0782}
\urldef\tempurl%
\url{https://doi.org/10.1145/2629489}
\showDOI{\tempurl}


\bibitem[{World Health Organization}(2022)]%
        {who2022variants}
\bibfield{author}{\bibinfo{person}{{World Health Organization}}.}
  \bibinfo{year}{2022}\natexlab{}.
\newblock \bibinfo{title}{Tracking SARS-CoV-2 variants}.
\newblock
  \bibinfo{howpublished}{\url{https://www.who.int/activities/tracking-SARS-CoV-2-variants}}.
\newblock


\bibitem[Wu et~al\mbox{.}(2020)]%
        {wu2020blink}
\bibfield{author}{\bibinfo{person}{Ledell Wu}, \bibinfo{person}{Fabio Petroni},
  \bibinfo{person}{Martin Josifoski}, \bibinfo{person}{Sebastian Riedel}, {and}
  \bibinfo{person}{Luke Zettlemoyer}.} \bibinfo{year}{2020}\natexlab{}.
\newblock \showarticletitle{Scalable Zero-shot Entity Linking with Dense Entity
  Retrieval}. In \bibinfo{booktitle}{\emph{Proceedings of the 2020 Conference
  on Empirical Methods in Natural Language Processing (EMNLP)}}.
  \bibinfo{publisher}{Association for Computational Linguistics},
  \bibinfo{address}{Online}, \bibinfo{pages}{6397--6407}.
\newblock
\urldef\tempurl%
\url{https://doi.org/10.18653/v1/2020.emnlp-main.519}
\showDOI{\tempurl}


\bibitem[Wu et~al\mbox{.}(2016)]%
        {wu2016features}
\bibfield{author}{\bibinfo{person}{Zhaohui Wu}, \bibinfo{person}{Yang Song},
  {and} \bibinfo{person}{C.~Lee Giles}.} \bibinfo{year}{2016}\natexlab{}.
\newblock \showarticletitle{Exploring Multiple Feature Spaces for Novel Entity
  Discovery}. In \bibinfo{booktitle}{\emph{Proceedings of the Thirtieth {AAAI}
  Conference on Artificial Intelligence}},
  \bibfield{editor}{\bibinfo{person}{Dale Schuurmans} {and}
  \bibinfo{person}{Michael~P. Wellman}} (Eds.). \bibinfo{publisher}{{AAAI}
  Press}, \bibinfo{address}{Phoenix, Arizona, {USA}},
  \bibinfo{pages}{3073--3079}.
\newblock
\urldef\tempurl%
\url{http://www.aaai.org/ocs/index.php/AAAI/AAAI16/paper/view/12261}
\showURL{%
\tempurl}


\bibitem[Xu and Miller(2022)]%
        {xu2022simple}
\bibfield{author}{\bibinfo{person}{Dongfang Xu} {and} \bibinfo{person}{Timothy
  Miller}.} \bibinfo{year}{2022}\natexlab{}.
\newblock \showarticletitle{A simple neural vector space model for medical
  concept normalization using concept embeddings}.
\newblock \bibinfo{journal}{\emph{Journal of Biomedical Informatics}}
  \bibinfo{volume}{130} (\bibinfo{year}{2022}), \bibinfo{pages}{104080}.
\newblock


\bibitem[Zhang et~al\mbox{.}(2020)]%
        {Zhang2020}
\bibfield{author}{\bibinfo{person}{Shuo Zhang}, \bibinfo{person}{Edgar Meij},
  \bibinfo{person}{Krisztian Balog}, {and} \bibinfo{person}{Ridho Reinanda}.}
  \bibinfo{year}{2020}\natexlab{}.
\newblock \showarticletitle{Novel Entity Discovery from Web Tables}. In
  \bibinfo{booktitle}{\emph{Proceedings of The Web Conference 2020}} (Taipei,
  Taiwan) \emph{(\bibinfo{series}{WWW '20})}. \bibinfo{publisher}{Association
  for Computing Machinery}, \bibinfo{address}{New York, NY, USA},
  \bibinfo{pages}{1298–1308}.
\newblock
\showISBNx{9781450370233}
\urldef\tempurl%
\url{https://doi.org/10.1145/3366423.3380205}
\showDOI{\tempurl}


\end{thebibliography}

\end{document}